\documentclass[10pt,twocolumn,letterpaper]{article}

\usepackage[pagenumbers]{cvpr} 

\usepackage[utf8]{inputenc} 
\usepackage[T1]{fontenc}    
\usepackage{url}            
\usepackage{booktabs}       
\usepackage{amsfonts}       
\usepackage{nicefrac}       
\usepackage{microtype}      

\usepackage{comment}
\usepackage{multirow}
\usepackage[table,xcdraw]{xcolor}
\usepackage{xspace}
\usepackage{makecell} 
\usepackage{amssymb}
\usepackage{pifont}
\usepackage{listings}
\usepackage{xcolor}
\usepackage{wrapfig}
\usepackage{tabularx}
\usepackage{siunitx}      
\usepackage{graphicx}      
\usepackage{subcaption}    
\usepackage[normalem]{ulem}  

\newcommand{\mypar}[1]{\vspace{1mm}\noindent\textbf{#1}}
\newcommand{\myparit}[1]{\vspace{1mm}\noindent\emph{#1}}

\def\eg{\emph{e.g.}\xspace}
\def\ie{\emph{i.e.}\xspace}

\definecolor{darkgray}{gray}{0.3}
\definecolor{darkgreen}{RGB}{50, 150, 50}
\definecolor{darkred}{RGB}{180, 50, 50}
\definecolor{darkblue}{RGB}{80, 106, 156}
\definecolor{col1}{HTML}{506A88}
\definecolor{col2}{HTML}{2E3D47}
\definecolor{col3}{HTML}{98A9B5}
\definecolor{col4}{HTML}{D19C83}
\definecolor{fom}{RGB}{0,153,139}
\definecolor{greenkeyword}{RGB}{34,139,34}
\definecolor{blueattribute}{RGB}{0,0,255}
\definecolor{lightgray}{RGB}{250,250,245} 

\usepackage[breakable, theorems, skins]{tcolorbox}

\usepackage{caption}
\usepackage{algorithm}
\usepackage{algorithmic}
\usepackage{longtable}
\usepackage{booktabs}
\usepackage{pifont}
\definecolor{codegray}{gray}{0.95}   
\definecolor{codepink}{rgb}{1,0.97,0.97}  
\lstdefinestyle{pystyle}{
  language=Python,
  backgroundcolor=\color{codepink},
  basicstyle=\ttfamily\scriptsize,
  showstringspaces=false,
  numbers=none,
  frame=single,
  rulecolor=\color{codepink},        
  breaklines=true,
  xleftmargin=0pt,
  mathescape=true
}

\usepackage[table]{xcolor}
\definecolor{headerbg}{HTML}{EEF3FB}
\definecolor{subheadbg}{HTML}{F7FAFF}
\definecolor{groupbg}{HTML}{F3FBF5}
\definecolor{rowalt}{HTML}{FAFCFF}

\definecolor{cvprblue}{rgb}{0.21,0.49,0.74}
\usepackage[pagebackref,breaklinks,colorlinks,allcolors=cvprblue]{hyperref}

\def\bench{\textsc{ConverSeg}\xspace}
\def\model{\textsc{ConverSeg-Net}\xspace}
\def\method{\textsc{ConverSeg-Net}\xspace}
\def\paperID{1561} 
\def\confName{CVPR}
\def\confYear{2026}

\title{Conversational Image Segmentation: \\Grounding Abstract Concepts with Scalable Supervision}

\author{Aadarsh Sahoo\\
California Institute of Technology\\
\and
Georgia Gkioxari\\
California Institute of Technology\\
}

\begin{document}

\twocolumn[{%
\renewcommand\twocolumn[1][]{#1}%
\maketitle
\begin{center}
    \centering
    \vspace{-7mm}
    \includegraphics[width=0.9\linewidth]{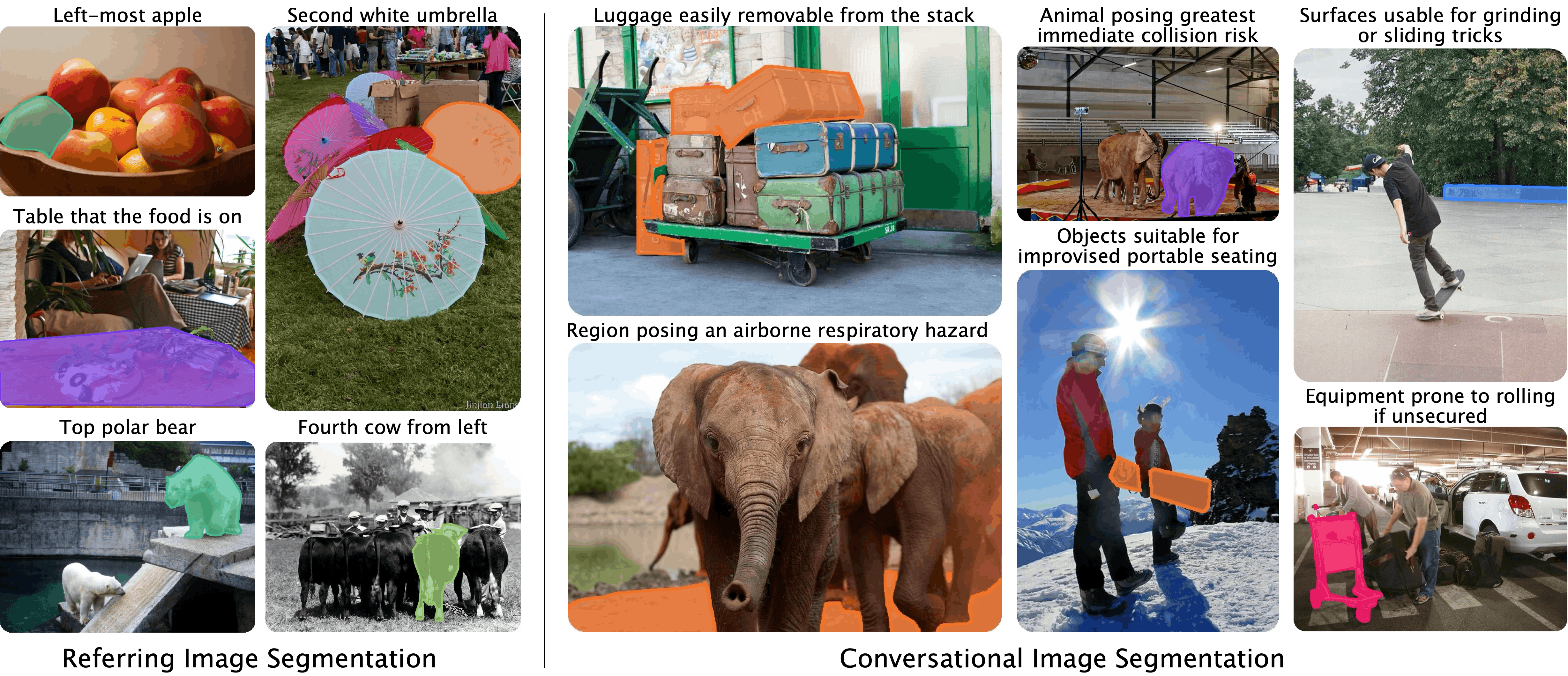}
    \vspace{-2mm}
    \captionof{figure}{\textbf{Conversational Image Segmentation requires reasoning beyond object categories.} \emph{Left}: Referring Image Segmentation (RIS) grounds descriptive phrases about object identity and spatial relations (e.g., ``left-most apple''). \emph{Right}: Conversational Image Segmentation (CIS) grounds abstract, intent-oriented concepts that require relational reasoning, physical understanding, and implicit constraints. 
    }
    \label{fig:teaser}
\end{center}
}]

\begin{abstract}

Conversational image segmentation grounds abstract, intent-driven concepts into pixel-accurate masks. Prior work on referring image grounding focuses on categorical and spatial queries (\eg, “left-most apple”) and overlooks functional and physical reasoning (\eg, “where can I safely store the knife?”). We address this gap and introduce Conversational Image Segmentation (CIS) and \bench, a benchmark spanning entities, spatial relations, intent, affordances, functions, safety, and physical reasoning. We also present \model, which fuses strong segmentation priors with language understanding, and an AI-powered data engine that generates prompt–mask pairs without human supervision. We show that current language-guided segmentation models are inadequate for CIS, while \model trained on our data engine achieves significant gains on \bench and maintains strong performance on existing language-guided segmentation benchmarks. Project webpage: \href{https://glab-caltech.github.io/converseg/}{https://glab-caltech.github.io/converseg}.

\end{abstract}

\section{Introduction}
\label{sec:intro}

\emph{``Which suitcases can I take without disturbing the stack?''} 
For humans, the answer is immediate: we identify which pieces are load-bearing versus accessible, anticipate redistribution of weight, and filter candidates by the constraint of ``easy removal.'' Yet a segmentation model, trained to predict \emph{suitcase} and \emph{cart}, lacks any representation of support relations, occlusion ordering, or physical stability. Selecting easily removable luggage requires reasoning jointly over geometry, physics, and user intent -- not merely recognizing object categories. This type of conversational, intent-driven language instruction reflects how humans naturally interact with their environments, yet remains beyond the reach of current perception systems.

In computer vision, grounding images using natural language expressions was first explored through the task of \emph{Referring Image Segmentation} (RIS)~\cite{yu2016modeling}. However, existing benchmarks for this task, RefCOCO variants~\cite{yu2016modeling}, primarily emphasize categorical and spatial references, such as “the white umbrella” or “the left-most apple”, shown on the left of~\cref{fig:teaser}.
By contrast, functional or physical reasoning about objects and environments -- such as “what object is prone to rolling if unsecured” or “where can I safely store the knife?” -- is largely underrepresented.

We address this gap by introducing \emph{Conversational Image Segmentation (CIS)}, a task that grounds high-level conversational concepts into pixel-accurate masks in natural images.
Examples are shown in~\cref{fig:teaser}.
We call these concepts conversational because they mirror how humans naturally talk about objects and their surroundings. 
They span five families, inspired by human vision science~\cite{gibson2014ecological, marr2010vision} and intuitive physics~\cite{battaglia2013simulation}:
(1)~\textbf{Entities} with open-vocabulary descriptions (\emph{“weathered wooden furniture”});
(2)~\textbf{Spatial \& Layout} capturing complex geometric relations (\emph{“items blocking the walkway”});
(3)~\textbf{Relations \& Events} describing interactions (\emph{“the player about to catch the ball”});
(4)~\textbf{Affordances \& Functions} requiring use-case reasoning (\emph{“surfaces suitable for hot cookware”}); and
(5)~\textbf{Physics \& Safety} involving stability or hazard assessment (\emph{“objects likely to tip over”}).

To measure progress in CIS, we introduce the \bench benchmark, featuring 1,687 human-verified image–mask pairs.  
Unlike previous benchmarks that focus on categorical entities and simple spatial relations, \bench offers coverage across all five concept families and a broader representation of conversational reasoning. 
We hope that advancing CIS with \bench will advance perception systems in assistive robotics, human–robot interaction and augmented reality -- domains that require grounding abstract concepts.

We further introduce \model, a conversational segmentation model that maps an image and prompt to a grounding mask. Training it demands large-scale supervision over diverse prompts in natural images -- a costly and cognitively intensive effort for human annotators who must produce pixel-accurate masks and natural, reasoning-rich prompts.
To bypass this bottleneck, we build an automated, VLM-driven data engine that synthesizes high-quality prompt–mask pairs \emph{without human supervision} via an iterative generate-and-verify loop, yielding 106K image–mask pairs across all five concept families. Trained on this data, \model achieves strong results on \bench and remains competitive on standard referring expression benchmarks, demonstrating both data quality and scalability.

In summary, our contributions are:
\begin{itemize}[leftmargin=*,nosep]
\item We introduce \emph{Conversational Image Segmentation (CIS)} and \bench, a benchmark of human-verified image–mask pairs targeting grounding of affordances, physics, and functional reasoning.

\item We build an AI-powered data engine that synthesizes diverse, high-quality conversational prompt–mask pairs without human supervision.

\item We design a baseline model, \model; trained on our engine’s data, it excels on \bench and remains strong on RIS benchmarks.

\end{itemize}

\section{Related Work}
\label{sec:related_work}

\begin{figure*}[t!]

\begin{center}

\includegraphics[width=\linewidth]{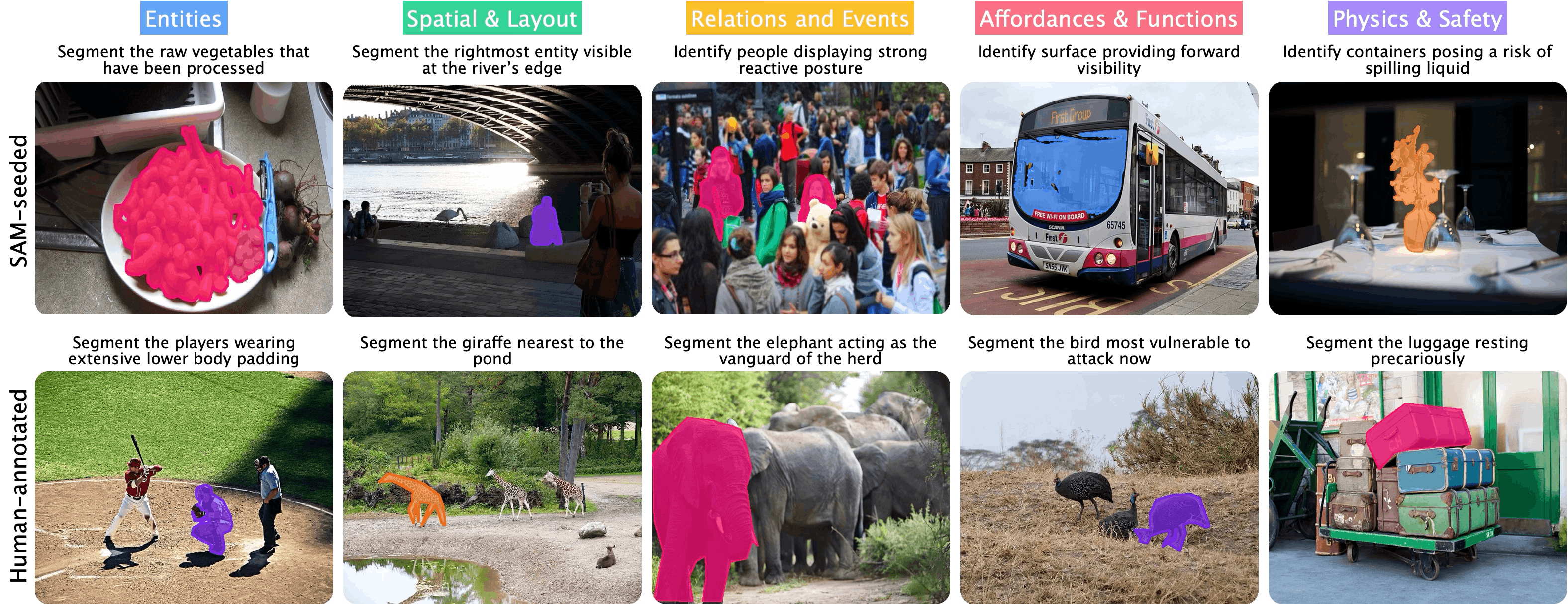} 

\end{center}

\vspace{-5mm}
\caption{
\small\textbf{\bench Qualitative Examples.} Representative samples across five concept categories. Prompts require reasoning about attributes, spatial relations, interactions, functional properties, and physical constraints beyond standard object reference.}
\label{fig:qualitative-benchmark}
\vspace{-6mm}
\end{figure*}

\mypar{Referring Expression Segmentation.}
Referring expression segmentation (RIS) localizes regions described by language. RefCOCO/+/g~\cite{yu2016modeling} are standard benchmarks but are dominated by object-centric, low-level spatial phrases (\eg, “person on the left,” “red cup”). Early methods used multi-stage language–vision pipelines~\cite{hu2016segmentation,liu2017recurrent}; recent work adopts Transformer-based vision-language encoders~\cite{yang2022lavt,kim2022restr}. Despite strong results on entities and simple spatial relations, these benchmarks seldom test affordances, stability, or user intent. Our CIS task and \bench explicitly target these gaps via five conversational concept families.

\mypar{Reasoning and Implicit Segmentation.}
ReasonSeg~\cite{lai2024lisa} pairs images with implicit, reasoning-heavy instructions and masks, but queries still target entities or spatial relations, with limited coverage of affordances, safety, or physical constraints. Multi-modal LLM systems (LISA~\cite{lai2024lisa}, GLaMM~\cite{rasheed2024glamm}, PixelLM~\cite{ren2024pixellm}) can perform multi-step reasoning and produce masks, yet rely on heavy backbones and multi-stage inference (chain-of-thought, tool calls), making deployment costly. In contrast, we pursue single-pass architectures that directly ground conversational concepts.

\mypar{Promptable Segmentation Models.}
The Segment Anything Model (SAM)~\cite{kirillov2023segment} enables promptable, class-agnostic segmentation from points or boxes; SAM2~\cite{ravi2024sam} extends this to streaming video. These offer strong priors but lack native text conditioning. Bridging this, some~\cite{ren2024grounded,liu2025segment} pair SAM with text-conditioned detectors, while others integrate SAM-like decoders into VLMs~\cite{rasheed2024glamm,zhang2024groundhog}. We leverage SAM’s learnt priors and combine them lightweight vision–language adapters to enable end-to-end conversational grounding without sacrificing segmentation quality.

\mypar{Vision-Language Models for Dense Prediction.}
Recent VLMs add heads for dense prediction: LISA~\cite{lai2024lisa} augments LLaVA~\cite{liu2023visual} with a mask decoder, GLaMM~\cite{rasheed2024glamm} supports multi-turn grounded dialogue, and GroundHog~\cite{zhang2024groundhog}, Kosmos-2~\cite{peng2023kosmos}, and xGen-MM/BLIP-3~\cite{xue2024xgen} push pixel-level grounding in ever-larger models. These systems excel at complex reasoning but demand substantial compute and often multiple passes per query. We take a complementary approach: rather than scaling model capacity, we scale \emph{training data diversity} through automated data synthesis. A lightweight 3B VLM + SAM2 decoder, trained on 106K auto-generated prompt–mask pairs across five reasoning concepts, achieves competitive CIS and RIS performance.

\mypar{Automated Data Synthesis.}
Synthetic data is a strong alternative to manual annotation, with synthetic pre-training approaching real-image performance for representation learning~\cite{sariyildiz2023fake}. Pipelines like ELEVATER~\cite{li2022elevater}, synthetic region captioning~\cite{you2023ferret, deng2025coconut}, and grounding verification~\cite{zhang2024groundhog} scale supervision, but mostly target literal descriptions or category labels. 
Our engine instead synthesizes conversational prompts aimed at affordances, layout constraints, and physical safety, then filters them via multi-stage visual verification.

\begin{figure}[t!]

\begin{center}
\includegraphics[width=\columnwidth]{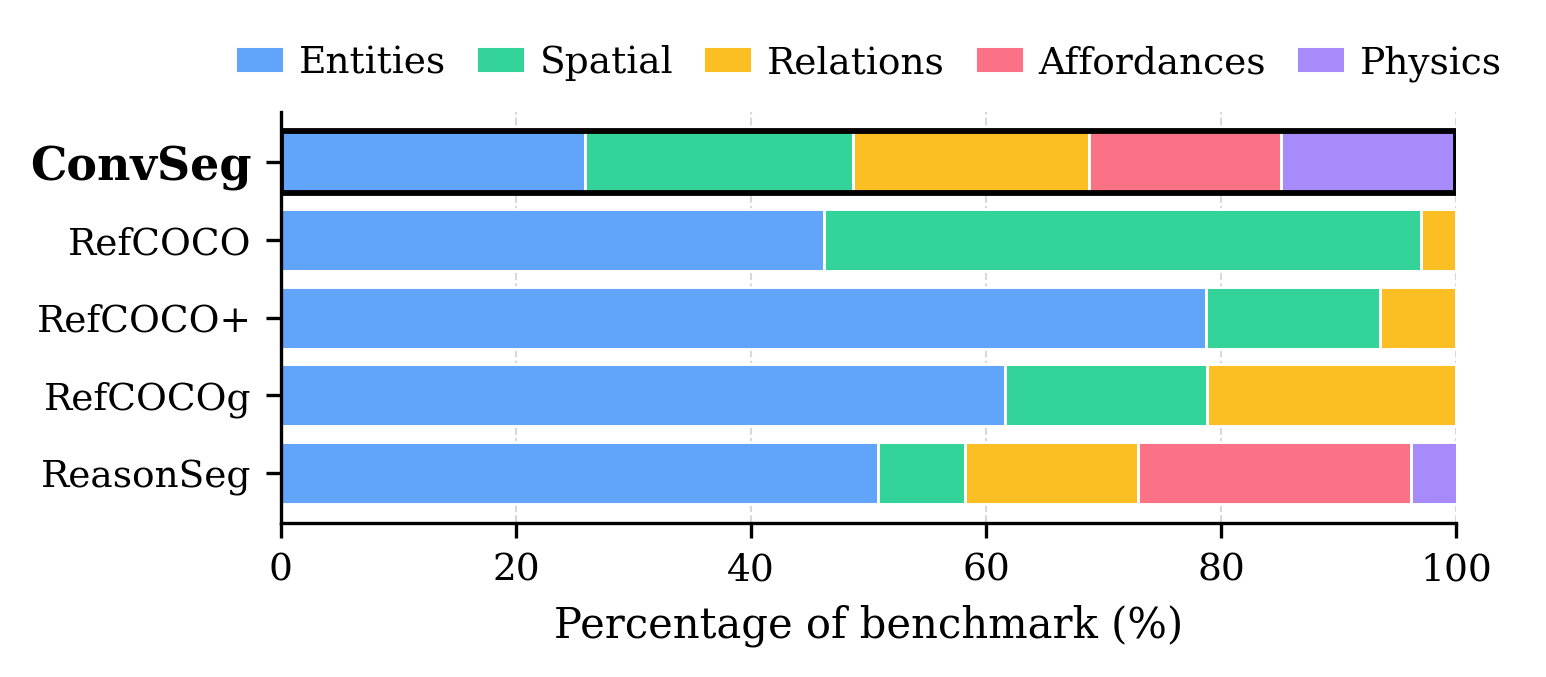} 
\end{center}

\vspace{-6mm}
\caption{
\small\textbf{Concept Coverage in Benchmarks.} Distribution of concepts across five existing benchmarks versus \bench. Prior datasets primarily focus on entities/spatial relations, whereas \bench offers near-uniform coverage across all five concepts.}
\label{fig:concept-distribution}
\vspace{-4mm}

\end{figure}
\begin{figure*}[t!]

\begin{center}

\includegraphics[width=0.96\linewidth]{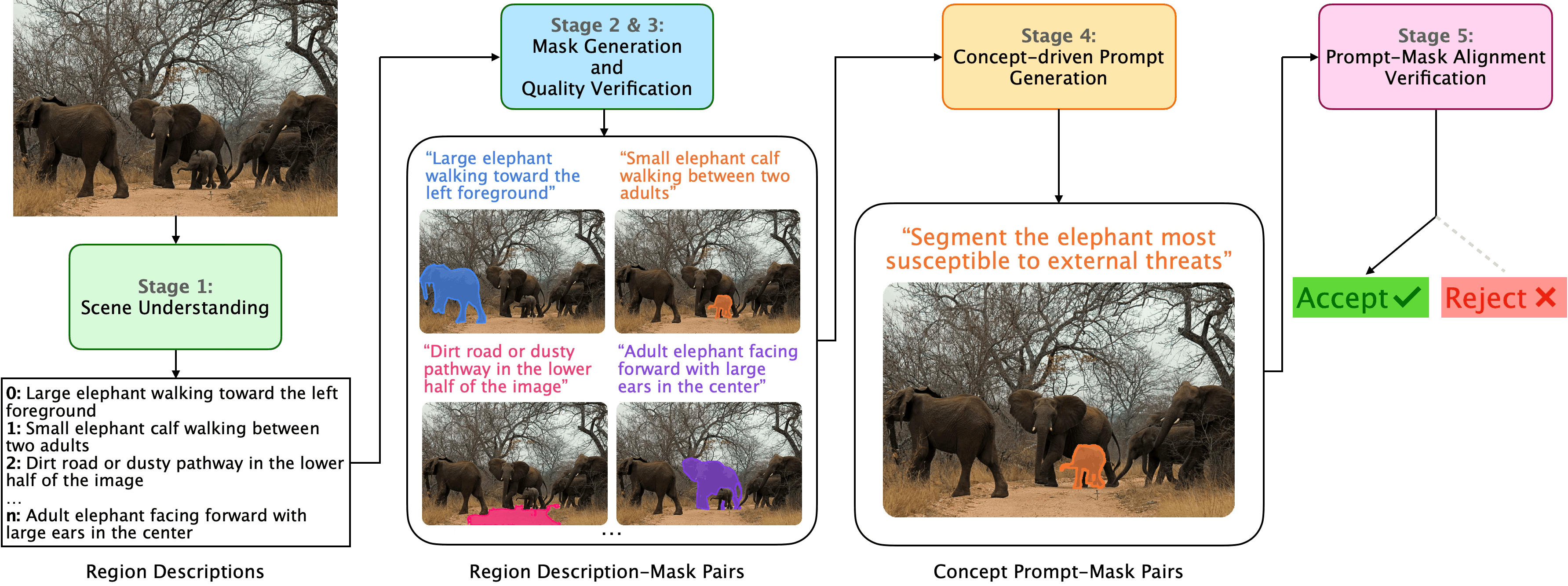}

\end{center}

\vspace{-6mm}
\caption{
\small\textbf{Conversational Data Engine.} Five-stage pipeline for automated concept-mask synthesis: (1) VLM generates region descriptions; (2) Object detector and SAM2 produce masks, VLM filters and refines them. (3) Concept-driven meta-prompts generate conversational queries assigned to region-mask pairs; (4) Alignment verification ensures prompt-mask correctness. 
}
\label{fig:dataengine}
\vspace{-2mm}
\end{figure*}

\section{Conversational Image Segmentation}
\label{sec:cis}

We introduce the task of \emph{Conversational Image Segmentation} (CIS) and the family of concepts that define it.

\subsection{Task Definition}
\label{sec:benchmark_definition}

Given an image $I$ and a natural language prompt $p$, the task is to predict a binary mask $M_p$ identifying the pixels in $I$ that satisfy the query $p$. Unlike referring image segmentation (RIS), prompts in CIS may require functional or physical reasoning (e.g., \emph{``surfaces stable enough to stack books''}), target non-visible properties such as affordances or safety (e.g., \emph{``objects that might be hot''}), and have context-dependent groundings (e.g., \emph{``comfortable seating areas''}).

\subsection{The Five Conversational Concepts}
\label{sec:benchmark_concepts}

We organize prompts into five concept families according to the type of reasoning they require. These families reflect how humans naturally query their surroundings and reveal reasoning abilities that extend beyond object-centric reference. We draw inspiration from human vision science~\cite{gibson2014ecological, marr2010vision} and intuitive physics~\cite{battaglia2013simulation}, which demonstrate that people infer functional properties and physical constraints directly from visual input -- capabilities largely missing from current vision tasks and benchmarks.

\begin{enumerate}
    \item \textbf{Entities.} Prompts that identify entities by category or attributes (e.g., \emph{``the bicycle with a basket''}). While overlapping with RIS, we include open-vocabulary categories and complex attribute compositions.
    
    \item \textbf{Spatial \& Layout.} Prompts about spatial relations, ordering, and occupancy (e.g., \emph{``the rightmost orange in the bowl''}, \emph{``the lamp behind the sofa''}).
    
    \item \textbf{Relations \& Events.} Prompts targeting interactions or transient states (e.g., \emph{``the player serving the ball''}, \emph{``the door being opened''}).
    
    \item \textbf{Affordances \& Functions.} Prompts requiring functional reasoning about object use (e.g., \emph{``surfaces you could cut on''}, \emph{``items that could serve as a shovel''}).
    
    \item \textbf{Physics \& Safety.} Prompts invoking stability, support, and hazard assessment (e.g., \emph{``objects likely to tip''}, \emph{``sharp objects posing a hazard''}).
\end{enumerate}

\section{The \bench Benchmark}
\label{sec:benchmark}

In this section, we introduce the \bench benchmark for conversational image segmentation and describe our data collection and annotation process.

~\cref{fig:concept-distribution} compares concept distributions across benchmarks. Existing datasets are heavily skewed toward entities and spatial relations ($>\!50\%$), with minimal coverage of affordances, events, or physical reasoning. In contrast, \bench provides balanced representation across all five concepts.

\subsection{Benchmark Construction and Verification}
\label{sec:benchmark_construction}

\bench is constructed through a two-stage pipeline: automated candidate generation followed by human verification. We source images from the COCO~\cite{lin2014microsoft} validation set and generate candidate tuples $(I, p, M_p, c)$ where $c \in \mathcal{C}$ denotes one of five concept types. 

\mypar{Segmentation masks} $M_p$ in \bench are either machine or human drawn, which lead to two evaluation sets:

\myparit{Human-annotated Set.}
We initialize masks with human-drawn annotations from COCO, sourced from instance (objects) and panoptic annotations (objects and stuff). This split has high-quality mask annotations but is restricted to annotations within COCO.

\myparit{SAM-seeded Set.} 
Masks are extracted with SAM2~\cite{ravi2024sam} prompted with bounding boxes generated by an object detector. This mode eliminates reliance on closed-vocabulary annotations from COCO and allows us to scale our evaluation data with minimal human supervision.

\mypar{Conversational prompts} $p$ are collected with the help of our data engine, described in detail in Sec.~\ref{sec:dataengine}, which proposes prompts $p$ with corresponding masks $M_p$ across all five concept families via a generate-then-verify process.

\mypar{Human Verification.} Each sample undergoes human verification for (i)~prompt quality and correct concept assignment and (ii)~mask accuracy. Verifiers accept/reject with a single click. The engine supplies image-tailored, diverse prompts across all five concept families; human checks ensure prompt and mask quality.

\subsection{\bench Statistics.}
\bench\ comprises 1,687 total samples across two splits: 1,194 SAM-seeded and 493 human-annotated. Prompts average 7.6 words (std: 1.2). \cref{fig:qualitative-benchmark} shows representative examples from each concept category. Additional statistics, annotation protocols, and qualitative examples are provided in the Appendix.

\section{The Conversational Data Engine}
\label{sec:dataengine}

We introduce a fully automatic data engine for conversational image segmentation. Collecting pixel-accurate masks and, more critically, realistic prompts is prohibitively expensive at scale with human annotators. Our data engine synthesizes high-quality prompt–mask pairs \textit{without human supervision}.
Instead, it leverages high-performing VLMs that iteratively generate outputs and verify their quality.

\subsection{Data Engine Architecture}
\cref{fig:dataengine} shows our pipeline, featuring modules for scene recognition, mask generation, prompt creation and verification.

\mypar{Stage 1: Scene Understanding.}
Given an image $I$, a VLM generates 
5–7 region descriptions in natural language, $\{d_1, \ldots, d_n\}$. 
Each $d_i$ specifies category, attributes, location, and relations 
in $\leq$15 words, \eg, \emph{``large elephant walking toward left foreground''}. These descriptions serve as targets for mask generation.

\mypar{Stage 2: Mask Generation.}
For each description $d_i$, we localize its segment in image $I$: Moondream3~\cite{moondream3} predicts a box $b_i$ from $(I,d_i)$; SAM2, prompted with $b_i$, returns the mask $m_i$. We choose Moondream3 for open-vocabulary detection and SAM2 for box-conditioned segmentation.

\mypar{Stage 3: Mask Quality Verification.}
Pipeline systems suffer from error propagation, so we enforce two checks for mask fidelity:

\myparit{Mask–text consistency check.} 
A VLM verifies that $(b_i, m_i)$ matches $d_i$ in identity, attributes, and spatial location. It returns accept/reject; only accepts proceed to the next stage.

\myparit{Mask refinement and selection.} 
Passing masks may still have under/over-coverage or holes, often from noisy boxes. We sample SAM2 with a dense point grid to form candidates and pick $m_i'$ with highest IoU to $m_i$. A VLM then chooses the better of the two based on coverage, boundary precision, artifacts. The selected ($\hat{m}_i$) is the final mask.

Now, each image $I$ has verified descriptions, $d_i$, paired with high-quality masks, $\hat{m}_i$.

\mypar{Stage 4: Concept-Driven Prompt Generation.}
This stage derives image-tailored conversational prompts for $I$ and its discovered regions.

Prompts must span a range of reasoning types within each concept $c$. For example, for affordances they should assess functional properties and context-dependent use ("surfaces safe for hot items"), canonical functions ("sources of water"), or counterfactual uses ("items that could prop a door"); for spatial relations, containment ("items inside containers") or ordinality ("the three leftmost cups"). We enforce this via concept-specific meta-prompts $\pi_c$ (found in the Appendix).

For each $c$, a VLM receives: (1) indexed descriptions $\{d_i\}$, (ii) set-of-marks numbered overlay~\cite{yang2023set}, and (iii) meta-prompt $\pi_c$. It generates up to three prompts $p$ per concept and selects the corresponding regions. Trivial prompt-mask pairs (\eg, an image with one car and “segment the car”) or prompts applicable to undiscovered regions are pruned.

\mypar{Stage 5: Prompt–Mask Alignment Verification.}
For each tuple $(I, p, M_p, c)$, a VLM verifies that $M_p$: (1) matches $p$’s target, (2) excludes irrelevant content, and (3) is reasonably described by $p$. It returns accept/reject; only accepts proceed.

This five-stage pipeline combines VLM-driven generation with multi-stage verification to produce high-quality prompt–mask pairs without human annotation. All VLM components use Gemini-2.5-Flash~\cite{comanici2025gemini}.

\mypar{Negative Data.} Beyond the positives, we also generate concept-specific negative prompts to improve robustness against plausible hallucinations. For each concept category, a VLM creates adversarial prompts using dedicated meta-prompts (see Appendix) that employ two strategies: (1) \textit{object-level neighbors} reference contextually plausible but absent objects (e.g., "segment the wine glass" at a dinner table setting without one), and (2) \textit{concept-level neighbors} describe present objects with incorrect attributes (e.g., "segment the wooden chair" for a metal one, or "segment the person standing" when they are sitting). Each negative prompt is verified by a VLM (Gemini-2.5-Flash) to ensure no valid mask exists in the image.

\subsection{Dual-Purpose Design}

Our data engine is crafted to generate diverse conversational prompt-mask pairs in images. 
We use it for two purposes:
\mypar{(1) Benchmark curation }(\cref{sec:benchmark_construction}). We run the pipeline on COCO val images, seeding mask candidates from COCO annotations or SAM2. For benchmark-grade quality, generated prompt–mask pairs receive a final human check.

\mypar{(2) Training data synthesis } (\cref{sec:model_training}). We run the data engine at scale on COCO train and SA-1B~\cite{kirillov2023segment} images, producing training pairs without human supervision. Ablations on pipeline components, failure mode analysis, and additional implementation details are in the Appendix.

\section{Model}
\label{sec:model}

In this section, we present a single-pass model for conversational image segmentation that \textit{grounds conversational concepts into pixels}. 
Our design philosophy is to avoid multi-component workflows like iterative tool use or multi-turn refinement to build a strong baseline for CIS. 
To this end, we combine advances in promptable image segmentation, but lack text conditioning, with VLMs that integrate vision and language but do not perform segmentation.
We describe our model architecture (\cref{sec:model_arch}), training strategy and implementation details (\cref{sec:model_training}).

\subsection{Model Architecture}
\label{sec:model_arch}

\begin{figure}[!t]

\begin{center}

\includegraphics[width=\linewidth]{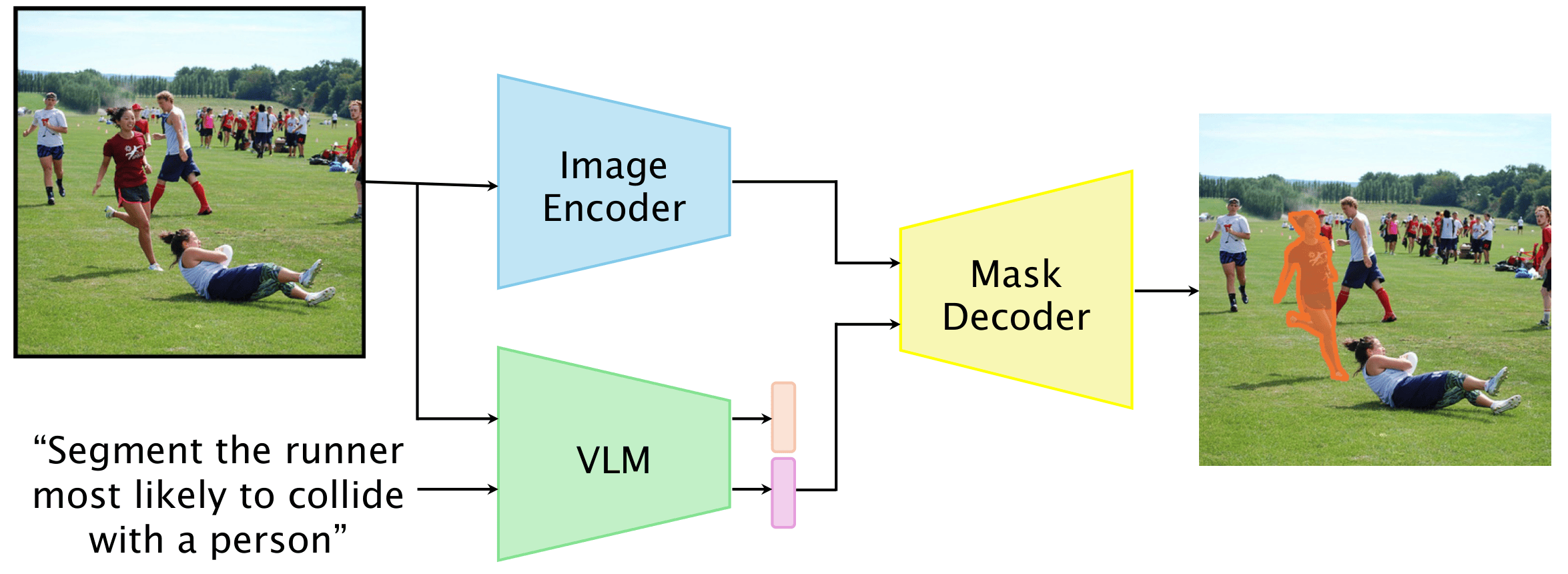} 
\end{center}

\vspace{-3mm}
\caption{\small\textbf{Model Architecture.} 
An image encoder processes the image, while a VLM jointly encodes image and text. Lightweight adapters map the text-token embeddings from the VLM to a mask decoder that predicts the target segment. 
}
\label{fig:model}
\vspace{-6mm}
\end{figure}

\cref{fig:model} illustrates our architecture. It fuses SAM2~\cite{ravi2024sam} (image-only components) with a compact vision–language backbone (Qwen-2.5-VL-3B~\cite{bai2025qwen2}), connected via lightweight prompt adapters.

\mypar{Image Encoder.}
We adopt the SAM2 image encoder which is an MAE~\cite{he2022masked} pre-trained Vision Transformer (ViT)~\cite{dosovitskiy2020image} adapted for high-resolution inputs and keep it frozen. The encoder processes each image once, independent of the prompt, producing a spatial image embedding $\mathbf{z}_{\text{img}} \in \mathbb{R}^{H' \times W' \times D_{\text{img}}}$.

\mypar{Prompt Encoder.}
We use Qwen2.5-VL-3B~\cite{bai2025qwen2} as the prompt encoder, which jointly processes the image $I$ and text prompt $p$ to produce per-token hidden states at the final layer. We extract hidden states corresponding to text tokens only (they have attended to the image tokens through Qwen's backbone), yielding a sequence $\{\mathbf{h}_1, \ldots, \mathbf{h}_T, \mathbf{h}_\texttt{EOS}\} \in \mathbb{R}^{D_t}$, where $T$ is the text length and $D_t$ is the hidden dimension.

Following SAM's design, we represent prompts as sparse and dense embeddings. The text token sequence $\{\mathbf{h}_1, \ldots, \mathbf{h}_T\}$ serves as sparse embeddings, capturing fine-grained text information. The hidden state at the \texttt{EOS} position serves as the dense embedding, capturing global image-text context. Two lightweight adapters project these representations to the decoder's input space: $\mathbf{e}_{\text{sparse}} = \text{Linear}_{D_t \to D_{\text{dec}}}(\{\mathbf{h}_1, \ldots, \mathbf{h}_T\})$ and $\mathbf{e}_{\text{dense}} = \text{MLP}_{D_t \to D_{\text{dec}}}(\mathbf{h}_{\texttt{EOS}})$, where the dense adapter is a 2-layer MLP with SiLU activation. The Qwen backbone is fine-tuned using LoRA~\cite{hu2022lora} with rank 16 and $\alpha=32$.

\mypar{Mask Decoder.}
We adopt SAM2's mask decoder and fully fine-tune it. The decoder uses modified Transformer blocks~\cite{vaswani2017attention} with bidirectional cross-attention between prompt and image embeddings. After two blocks, the image embedding is upsampled and an MLP maps the output token to per-pixel foreground probabilities, producing the final mask.

\subsection{Training Curriculum}
\label{sec:model_training}

To tackle the reasoning-heavy task of conversational image segmentation, we use a curriculum that gradually increases task complexity: the model first learns to segment literal concepts (\eg, ``segment the cat'') before advancing to more abstract referential and conversational concepts. This approach is critical because SAM2 has no prior exposure to language and thus requires careful integration
of text conditioning. We validate this design via ablations in \cref{sec:ablations}.

\subsubsection{Training Data}
\label{sec:training_data}

We organize training data into four groups with increasing complexity:

\mypar{(1) Literal concepts.} We leverage the COCO train set reformulated as category-level segmentation using refined masks from COCONut~\cite{deng2024coconut}. For each image, we randomly sample a category and treat all corresponding masks as ground truth with prompts like \emph{``Segment all the [category] in the image''}. In addition, we also get part-level segmentation masks from PACO~\cite{ramanathan2023paco}. This yields 440K prompt-mask pairs.

\mypar{(2) Basic referring expressions.} We utilize the  RefCOCO family datasets~\cite{yu2016modeling} (RefCOCO, RefCOCO+, RefCOCOg) train splits, providing 321K object-centric references.

\mypar{(3) Open-vocabulary regions.} Region descriptions from Stage 3 of our data engine (Sec.~\ref{sec:dataengine}), providing 48K diverse region descriptions beyond COCO's closed vocabulary. Examples are shown in~\cref{fig:dataengine}.

\mypar{(4) Conversational concepts.} This is the output from our data engine which generates 106K concept-mask pairs, covering all five concept families, plus an equal number of concept-specific negative prompts with empty masks to improve robustness.

\subsubsection{Two Phase Training} 

We train in two phases with increasing task complexity:

\mypar{Phase 1: Pretraining.} We pretrain on a mixture of groups 1-3, producing a base model proficient at basic referring segmentation.

\mypar{Phase 2: Conversational post-training.} Initialized from Phase 1, we fine-tune on group 4 mixed with samples randomly drawn from groups 1 through 3 such that positives, negatives, and pretraining data are equal in proportion.
This mixing strategy maintains performance on foundational segmentation tasks while adapting to conversational concepts, empirically validated in~\cref{sec:ablations}.

\subsubsection{Training Objective}

Given binary ground-truth mask $M \in \{0,1\}^{H \times W}$ and predicted probability mask $M^* \in [0,1]^{H \times W}$, we minimize a weighted combination of binary cross-entropy and Dice loss: $\mathcal{L} = \mathcal{L}_{\text{BCE}}(M, M^*) + \lambda \mathcal{L}_{\text{Dice}}(M, M^*)$ with $\lambda = 0.25$.

\subsubsection{Implementation Details}
We train with AdamW~\cite{loshchilov2017decoupled}, batch size 6 with gradient accumulation of 8 steps, and a cosine schedule with warmup. Phase 1 runs for 100K steps and Phase 2 runs for 90K steps. We use a learning rate of $\eta{=}1\text{e}{-4}$. Training on a single NVIDIA A100 80GB takes about 96 hours. Additional details are in the Appendix.

\section{Experiments}
\label{sec:experiments}

We evaluate our model against baselines on conversational image grounding, outlining the setup (\cref{sec:exp_setup}), results (\cref{sec:main_results}), and ablation studies and analysis (\cref{sec:ablations}).

\newcolumntype{A}{>{\columncolor{black!5}}c} 

\begin{table*}[!ht]
\centering
\scriptsize
\setlength{\tabcolsep}{3pt}
\renewcommand{\arraystretch}{1.12}
\resizebox{0.85\linewidth}{!}{
\begin{tabular}{l l A *{5}{c} A *{5}{c}}
\toprule
\multirow{2}{*}{Model} & \multirow{2}{*}{Prompt Encoder}
& \multicolumn{6}{c}{SAM-seeded (gIoU)} 
& \multicolumn{6}{c}{Human-annotated (gIoU)} \\
& & All & Ent. & Spat. & Rel. & Aff. & Phys. & All & Ent. & Spat. & Rel. & Aff. & Phys. \\
\midrule


LISA\textsuperscript{$\star$}                 & LLaVA 7B      & 48.6  & 52.3  & 55.0  &  54.5 & 41.5  & 39.1  &  45.9 & 46.5  & 50.4  & 50.9  & 42.6  &  37.7 \\
LISA\textsuperscript{$\star$}                 & Llama2 13B     &  55.2 &  60.0 &  57.1 & 60.3  & 50.1  &  46.6 & 53.8  & 54.3  &  60.1 & 56.8  &  53.6 &  42.1 \\
UniLSeg-20  & CLIP ViT-B/16    &  32.6 & 39.0  & 36.0  & 36.5  & 26.4  &  22.8 & 31.6  &  36.6 & 35.4  & 27.4  &  27.6 &  29.4 \\
EVF-SAM\textsuperscript{$\dagger$}  & BEIT-3-Large            & 42.2  & 47.1  & 49.3  & 46.0  & 37.3  & 29.8  &  39.1 & 42.1  & 50.1  & 41.3  & 29.5  & 31.2  \\
EVF-SAM\textsuperscript{$\ddagger$}   & BEIT-3-Large    & 47.7  &  53.8 &  54.7 & 50.9  & 40.9  & 36.6  &  45.4 &  51.9 & 52.2  &  44.2 &  41.4 &  34.1 \\
Seg-Zero            & Qwen2.5-VL 7B   &  69.2 &  74.1 &  71.7 &  72.3 &  65.1 & 60.9  & 61.1  & 62.8  & 63.6  & 61.1  &  60.2 & 56.6  \\
\midrule
\model (Base)     & Qwen2.5-VL 3B     &   58.0 & 66.0 & 60.5 & 64.6 & 52.3 & 41.8 & 56.5 & 61.9 & 59.9 & 59.4 & 52.7 & 45.5 \\
\model     & Qwen2.5-VL 3B     &   \textbf{70.8} & 74.0 & 70.9 & 74.1 & 68.7 & 64.2 & \textbf{67.4} & 71.6 & 68.7 & 67.0 & 64.4 & 63.8 \\
\model     & Qwen2.5-VL 7B     &   \textbf{72.4 }& 76.1 & 71.1 & 77.5 & 70.4 & 63.7 & \textbf{67.9} & 70.0 & 71.5 & 69.3 & 63.5 & 64.0 \\
\bottomrule
\end{tabular}
}
\vspace{-2mm}
\caption{\textbf{\bench benchmark results (gIoU, \%).}
Each subset reports performance across the five concept categories -- \uline{Ent}ities, \uline{Spat}ial, \uline{Rel}ations, \uline{Aff}ordances, and \uline{Phys}ics \& Safety -- and summarizes across all (\emph{All}). $^\star$ indicates models fine-tuned on ReasonSeg training data. $^\dagger$ trained on RefCOCO only; $^\ddagger$ trained on RefCOCO and additional datasets (Objects365, PASCAL-Part, etc).}
\vspace{-3mm}
\label{tab:convseg_results}
\end{table*}




\subsection{Experimental Setup}
\label{sec:exp_setup}

\mypar{Benchmarks.}
We evaluate on three benchmarks: (1) \bench: our benchmark with SAM-seeded and human-annotated splits across five concept categories (entities, spatial relations, affordances, relations \& events, physics \& safety); (2) RefCOCO/+/g~\cite{yu2016modeling}: standard referring expression benchmarks; (3) ReasonSeg~\cite{lai2024lisa}: segmentation benchmark of complex implicit reasoning and understanding. 

\mypar{Baselines.} We compare against state-of-the-art feed-forward methods of diverse architectural designs: LISA~\cite{lai2024lisa}, an MLLM-based model with embedding-as-mask paradigm; UniLSeg~\cite{liu2024universal}, a universal model for arbitrary semantic granularity; EVF-SAM~\cite{zhang2024evf}, a SAM-based model with early vision-language fusion; and Seg-Zero~\cite{liu2025seg}, a reasoning-chain guided framework with decoupled reasoning and segmentation modules. Qualitative comparisons with LISA appear in~\cref{fig:qualitative-predictions}; additional baselines are in the Appendix.

\mypar{Evaluation Metrics.}
We report generalized IoU (gIoU) as our primary metric following~\cite{kazemzadeh2014referitgame,yu2016modeling}. Cumulative IoU (cIoU) results are provided in the Appendix.

\subsection{Main Results}
\label{sec:main_results}

\begin{figure*}[!t]

\begin{center}

\includegraphics[width=0.95\linewidth]{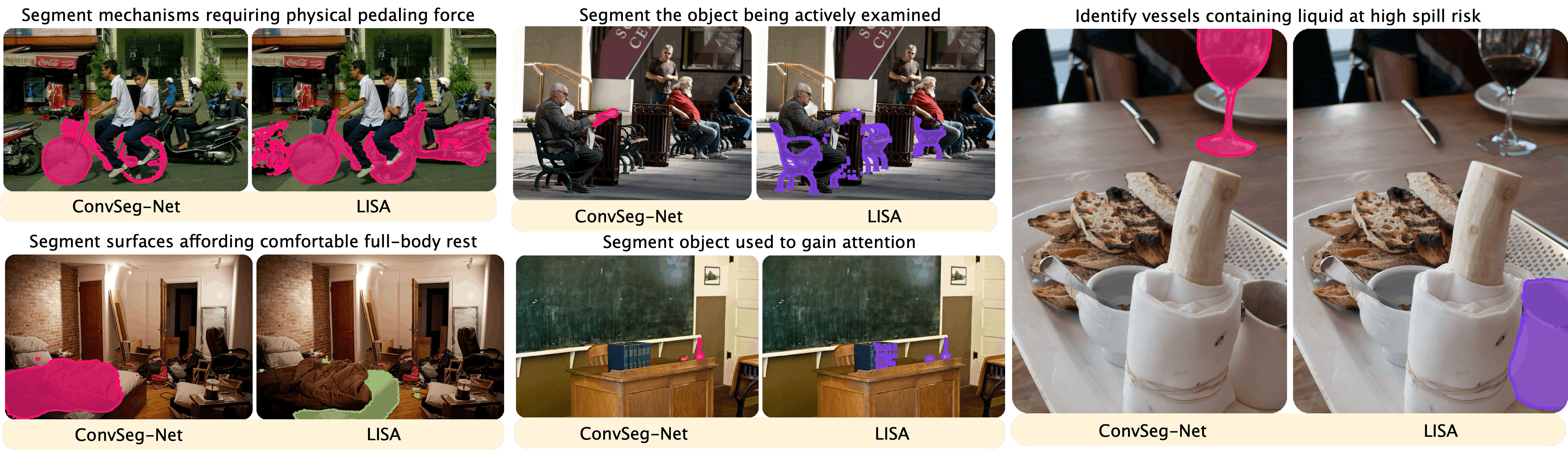} 

\end{center}

\vspace{-5mm}
\caption{Qualitative results on \bench. \model produces more accurate masks for spatial, affordance, and physics concepts than LISA, a popular leading reasoning segmentation model.
}
\label{fig:qualitative-predictions}
\vspace{-4mm}
\end{figure*}

\mypar{Results on \bench.}
\cref{tab:convseg_results} reports overall and per-concept gIoU on the SAM-seeded and human-annotated splits of \bench. Our Phase-1 model (\method Base, Qwen2.5-VL 3B, without conversational training) already achieves 58.0\% on the SAM-seeded split surpassing the strongest LISA variant (55.2\% with Llama2-13B) by +2.8\%, despite using a 4$\times$ smaller backbone and without ReasonSeg fine-tuning. Among existing methods, Seg-Zero is the strongest baseline with 69.2\% overall. Our full 3B model (\method) reaches 70.8\%, improving over Seg-Zero by +1.6\% on the SAM-seeded split. Scaling the prompt encoder to 7B further boosts performance to 72.4\%, a +3.2\% absolute gain over the best baseline. The same trends hold on the human-annotated split.

\mypar{Per-Concept Analysis.}
\cref{tab:convseg_results} reports per-concept performance. Key takeaways:
(1) baselines score highest on entities and spatial and lowest on affordances, physics \& safety concepts. For example, LISA-Llama2-13B on the SAM-seeded split achieves 60.0\% on entities and 46.6\% on physics \& safety (-13.4\%). (2) Our
base model shows an even larger gap: 66.0\% vs.\ 41.8\% (–24.2\%). (3) Phase-2 conversational training boosts performance for all concepts, with biggest gains for physics \& safety (from 41.8\% to 64.2\%) narrowing the the gap with entities to 9.8\% (74.0\% vs.\ 64.2\%). 
(4) Scaling the backbone (\method-7B) achieves the best results across all families.
Overall, later-stage training and larger models especially
benefit abstract concepts while preserving strong entity-level performance.

\mypar{Qualitative Examples.}
Fig.~\ref{fig:qualitative-predictions} compares \model and LISA predictions on \bench.
\model better localizes abstract concepts: it focuses on the bicycle as the mechanism requiring physical pedaling force, isolates the newspaper as the object being actively examined, selects only vessels containing liquid at high spill risk, and segments the bell as the object used to gain attention, whereas LISA often includes nearby distractors or misses subtle physical cues. For ``surfaces affording comfortable full-body rest,'' however, LISA’s decision to segment the full bed is arguably more canonical than \model’s blanket-focused mask.

\newcolumntype{A}{>{\columncolor{black!5}}c} 
\begin{table*}[t]
\centering
\footnotesize
\setlength{\tabcolsep}{4pt}
\renewcommand{\arraystretch}{1.15}
\resizebox{0.9\linewidth}{!}{
\begin{tabular}{l l A *{2}{c} A *{2}{c} A *{2}{c} c A c c}
\toprule
\multirow{2}{*}{Model} & \multirow{2}{*}{Prompt Encoder} & \multicolumn{3}{c}{RefCOCO} & \multicolumn{3}{c}{RefCOCO+} & \multicolumn{3}{c}{RefCOCOg} & \multicolumn{4}{c}{ReasonSeg} \\
\cmidrule(lr){3-5}\cmidrule(lr){6-8}\cmidrule(lr){9-11}\cmidrule(lr){12-15}
& & val & testA & testB & val & testA & testB & val(U) & test(U) & val(G) & val & test & test(short) & test(long) \\
\midrule

LISA      & LLaVA 7B      & 74.9 & 79.1 & 72.3 & 65.1 & 70.8 & 58.1 & 67.9 & 70.6 & -- & 44.4 & 36.8  & 37.6  & 36.6 \\
LISA$^\star$      & LLaVA 7B        & -- & -- & -- & -- & -- & -- & -- & -- & -- & 52.9 & 47.3  & 40.6  & 49.4 \\
LISA$^\star$      & Llama2 13B     & -- & -- & -- & -- & -- & -- & -- & -- & -- & 60.0 & 51.5  & 43.9  & 54.0 \\
Seg-Zero             & Qwen2.5-VL 3B   & -- & -- & -- & -- & -- & -- & -- & -- & -- & 58.2 & 56.1  & -- & -- \\
Seg-Zero             & Qwen2.5-VL 7B   & -- & -- & -- & -- & -- & -- & -- & -- & -- & \textbf{62.6} & 57.5  & -- & -- \\
GSVA$^\star$      & Vicuna 13B           & 79.2 & 81.7 & 77.1 & 70.3 & 73.8 & 63.6 & 75.7 & 77.0 & -- & -- & -- & -- & -- \\
GLaMM     & Vicuna 7B     & 79.5 & 83.2 & 76.9 & 72.6 & 78.7 & 64.6 & 74.2 & 74.9 & -- & 47.4 & --  & -- & -- \\
UniLSeg-20   & CLIP ViT-B/16    & 80.5 & 81.8 & 78.4 & 72.7 & 77.0 & 67.0 & 78.4 & 79.5 & -- & -- & -- & -- & -- \\
EVF-SAM$^\dagger$   & BEIT-3-Large            & 82.1 & 83.7 & 80.0 & 75.2 & 78.3 & 70.1 & 76.8 & 77.4 & -- & -- & -- & -- & -- \\
EVF-SAM$^\ddagger$   & BEIT-3-Large    & 82.4 & 84.2 & 80.2 & 76.5 & 80.0 & 71.9 & 78.2 & 78.3 & -- & -- & -- & -- & -- \\
HyperSeg  & Phi2 2.7B     & 84.8 & 85.7 & \textbf{83.4} & \textbf{79.0} & \textbf{83.5} & \textbf{75.2} & 79.4 & 78.9 & -- & -- & -- & -- & -- \\
HyperSeg  & Phi2 3B         & -- & -- & -- & -- & -- & -- & -- & -- & -- & 59.2 & -- & -- & -- \\
Gemini Seg & Gemini2.5 Flash & -- & -- & -- & -- & -- & -- & -- & -- & -- & 28.3 & 30.6  & 16.5  & 35.0 \\
X-SAM$^\star$    & Phi3 3.8B     & \textbf{85.1} & \textbf{87.1} & \textbf{83.4} & 78.0 & 81.0 & 74.4 & \textbf{83.8} & \textbf{83.9} & -- & 56.6 & \textbf{57.8}  & 47.7  & 56.0 \\
RSVP       & LLaVA1.6 7B     & -- & -- & -- & -- & -- & -- & -- & -- & -- & 59.2 & 56.9  & 47.9  & \textbf{58.4} \\
RSVP       & Qwen2-VL 7B     & -- & -- & -- & -- & -- & -- & -- & -- & -- & 58.6 & 56.1  & 48.5  & 57.1 \\

\midrule
\model (Base) & Qwen2.5-VL 3B     &  78.4 & 80.8 & 75.8 & 72.5 & 77.7 & 66.4 & 75.1 & 74.7 & 74.7 & 51.1 & 48.3 & 47.2 & 48.6 \\
\model & Qwen2.5-VL 3B     &  78.2 & 80.3 & 74.7 & 72.0 & 77.5 & 66.3 & 74.1 & 73.9 & 73.7 & 56.4 & 52.2 & 53.8 & 51.7 \\
\model & Qwen2.5-VL 7B     &  79.4 & 81.6 & 76.4 & 74.3 & 79.1 & 69.2 & 74.9 & 75.5 & 75.0 & 61.9 & 57.0 & \textbf{54.2} & 57.9 \\

\bottomrule
\end{tabular}
}
\vspace{-2mm}
\caption{\textbf{Referring expression segmentation (gIoU, \%).} \method is competitive on RefCOCO/+/g and achieves strong zero-shot performance on ReasonSeg with the 7B model, surpassing some methods fine-tuned on ReasonSeg ($^\star$). $^\dagger$ trained on RefCOCO only; $^\ddagger$ on RefCOCO plus additional datasets (Objects365, PASCAL-Part, etc).}

\vspace{-4mm}
\label{tab:refebench-results}
\end{table*}

\mypar{Comparison on Referring Expression Benchmarks.}
\cref{tab:refebench-results} reports gIoU on RefCOCO/+/g~\cite{yu2016modeling} and ReasonSeg~\cite{lai2024lisa}. 
While the primary goal of \method is to establish a strong baseline for conversational image segmentation with abstract concepts, we evaluate on standard referring expression benchmarks to demonstrate robustness on literal concepts.
On RefCOCO, \method~achieves 78.4\% on val, competitive with models like GSVA (79.2\%) and EVF-SAM (82.4\%) that use substantially more training data. 
We note that RefCOCO/+/g datasets contain noisy ground truth annotations with incorrect or inaccurate masks, which can artificially lower performance metrics even when model predictions are reasonable. We provide qualitative examples of such cases in the Appendix.
On ReasonSeg, our 3B model reaches 52.2\% on the test set without training on any ReasonSeg data, a few points ahead of LISA-13B (51.5\%), which was fine-tuned on its training set and is 4$\times$ larger. Scaling our model to 7B improves performance (57.0\% on ReasonSeg test), showing that our conversational training effectively transfers to complex reasoning scenarios zero-shot (\ie, without task-specific supervision).

\subsection{Ablation Studies and Analysis}
\label{sec:ablations}

\begin{table*}[t]
  \centering
  \small
  \setlength{\tabcolsep}{6pt}  

  \begin{minipage}[t]{0.40\textwidth}
  \vspace{0pt}
    \centering
    \resizebox{\linewidth}{!}{
      \begin{tabular}{lcc}
        \toprule
        Training Strategy & RefCOCO/+/g & \bench \\
        \midrule
        No curriculum (only conversational)      &     56.1      & 66.0 \\
        No curriculum (all data)                &      \textbf{75.5}     & 65.4 \\
        Phase 1 + Phase 2 (only conversational) &      57.7     & 65.2 \\
        \midrule
        Phase 1                                &      75.1     &  56.5 \\
        Phase 1 + Phase 2 (full curriculum)    & 74.5 & \textbf{67.4} \\
        \bottomrule
      \end{tabular}
    }
    \vspace{-2mm}
    \caption{Curriculum learning ablation on RefCOCO/+/g and \bench. 
    }
    \label{tab:ablation_curriculum}
  \end{minipage}
  \hfill
  \begin{minipage}[t]{0.32\textwidth}
  \vspace{0pt}
    \centering
    \resizebox{\linewidth}{!}{
      \begin{tabular}{lcc}
        \toprule
        Architecture Configuration & \bench & $\Delta$ \\
        \midrule
        w/o LoRA finetuning     & 48.3 & -19.1 \\
        Text-only input to Qwen & 49.5 & -17.9 \\
        Sparse embeddings only  & 67.3 & -0.1 \\
        \midrule
        \model                  & \textbf{67.4} & -- \\
        \bottomrule
      \end{tabular}
    }
    \vspace{-2mm}
    \caption{Architectural ablations. 
    Each row removes one component from the final design.
    }
    \label{tab:ablation_architecture}
  \end{minipage}
  \hfill
  \begin{minipage}[t]{0.25\textwidth}
  \vspace{0pt}
    \centering
    \resizebox{\linewidth}{!}{
      \begin{tabular}{lc}
        \toprule
        Prompt Encoder      & \bench \\
        \midrule
        Perception-LM-3B    & 66.5 \\
        Qwen2.5-VL-3B       & 67.4 \\
        \bottomrule
      \end{tabular}
    }
    \vspace{-2mm}
    \caption{Prompt encoder backbone comparison on \bench. 
    }
    \label{tab:ablation_perception-lm}
  \end{minipage}

  \vspace{-4mm}
\end{table*}

\mypar{Curriculum Learning.}
\cref{tab:ablation_curriculum} analyzes our training curriculum on RefCOCO/+/g (avg. over 9 splits) and \bench's human-annotated split. Training only on conversational data yields reasonable \bench performance (66.0\%) but poor RefCOCO/+/g results (56.1\%), indicating overfitting. Jointly training on all data without a curriculum improves RefCOCO/+/g to 75.5\% but degrades \bench to 65.4\%. A two-phase curriculum (Phase 1: basic referring data; Phase 2: conversational data) achieves 57.7\% and 65.2\%. Our final strategy -- Phase 1 on basic data followed by Phase 2 with a mix of both data types -- achieves the highest overall performance (74.5\% on RefCOCO/+/g, 67.4\% on \bench), showing that curriculum learning with mixed fine-tuning effectively balances performance across benchmarks.

\mypar{Architectural Ablations.}
\cref{tab:ablation_architecture} ablates key architectural choices. 
Freezing the prompt encoder backbone instead of using LoRA fine-tuning causes a drastic performance drop (48.3\% vs. 67.4\%) -- adapting the prompt encoder is crucial for language grounding. Providing only text input to the prompt encoder, rather than both image and text, reduces performance by 17.9\% (67.4\% to 49.5\%), showing that visual context is essential for text-conditioned segmentation. Finally, removing dense prompt embeddings while retaining only sparse embeddings causes a modest 0.1\% drop.

\mypar{VLM Backbone Comparison.}
\cref{tab:ablation_perception-lm} evaluates Perception-LM-3B~\cite{cho2025perceptionlm}, a recent VLM with strong visual reasoning capabilities. Performance remains comparable (66.5 vs. 67.4), demonstrating that \model can work with any performant VLM backbone.

\begin{figure}[!thbp]

\begin{center}
\includegraphics[width=\columnwidth]{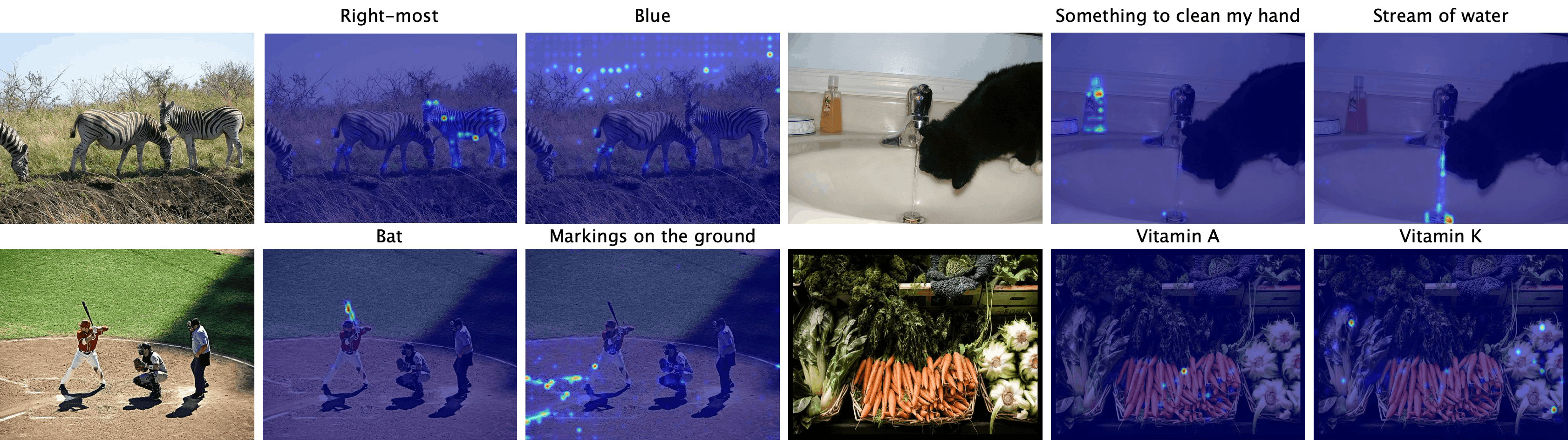} 
\end{center}

\vspace{-4mm}
\caption{
\small Cross-attention maps showing that each text prompt focuses on its corresponding image region. 
}
\label{fig:attention-maps}
\vspace{-6mm}

\end{figure}

\mypar{Attention Map Visualization.}
\cref{fig:attention-maps} shows cross-attention between text tokens and image regions in the mask decoder. For each image we visualize maps for two prompts; attention concentrates on the referred region (\eg ``Vitamin A'' on carrots) and is sparse and point-like rather than diffuse. We hypothesize this stems from conditioning the SAM mask decoder with language embeddings in place of its point embeddings, so each token behaves like a soft point prompt.

\section{Conclusion}
\label{sec:conclusion}

We introduced Conversational Image Segmentation (CIS), grounding high-level concepts about affordances, physics, and function into pixel-accurate masks. Our \bench benchmark provides 1,687 human-verified samples with balanced coverage across five concept families underrepresented in prior work. To scale beyond manual annotation, we built an automated data engine synthesizing 106K prompt-mask pairs via iterative VLM generation and verification. \model, trained on this data, achieves state-of-the-art performance on \bench while remaining competitive on standard benchmarks, showing that a curriculum from literal to conversational concepts effectively adapts promptable segmentation to language conditioning.

\section*{Acknowledgments}
We thank Damiano Marsili and Ilona Demler for their valuable feedback.
Aadarsh is supported by the Kortschak Scholarship and Caltech's CAST program. Georgia is supported by the Powell Foundation, Meta through the LLM evaluation research grant, Google, and Amazon.

{
    \small
    \bibliographystyle{ieeenat_fullname}
    \bibliography{main}
}

\newpage

\clearpage
\appendix



\newcommand{\tocline}[3]{
  \noindent#1#2\dotfill\ #3\par\vspace{0.15em}%
}
\newcommand{\tocsection}[2]{\tocline{}{\textbf{#1}}{\pageref{#2}}}
\newcommand{\tocsubsec}[2]{\tocline{\hspace{1.8em}}{#1}{\pageref{#2}}}
  {\Large\bfseries Appendix}
\vspace{0.8em}

\noindent
\begin{minipage}[t]{0.48\textwidth}
\tocsection{A\quad Qualitative Results}{sec:qualitative}
  \tocsubsec{A.1\quad Qualitative Predictions by \model}{sec:qualitative:additional-predictions}
  \tocsubsec{A.2\quad Failure Cases}{sec:qualitative:failure-cases}
  \tocsubsec{A.3\quad Annotation Quality in RefCOCO/+/g}{sec:qualitative:refcoco-quality}

\tocsection{B\quad Conversational Data Engine Details}{sec:data_engine}
  \tocsubsec{B.1\quad Meta-Prompts and Stage Details}{sec:data_engine:meta_prompts}
  \tocsubsec{B.2\quad Negative Data Generation}{sec:data_engine:negative_data}

\tocsection{C\quad Benchmark Construction and Analysis}{sec:benchmark}
  \tocsubsec{C.1\quad Annotation Protocols}{sec:benchmark:annotation}
  \tocsubsec{C.2\quad VLM Verifier Reliability for \bench}{sec:benchmark:ai-verifier}
  \tocsubsec{C.3\quad Additional Statistics and Visualizations}{sec:benchmark:stats}
  \tocsubsec{C.4\quad Additional Qualitative Examples}{sec:benchmark:additional-benchmark}
\end{minipage}\hfill
\begin{minipage}[t]{0.48\textwidth}
\tocsection{D\quad Implementation Details}{sec:implementation}
  \tocsubsec{D.1\quad Architecture}{sec:implementation:architecture}
  \tocsubsec{D.2\quad Training Hyperparameters}{sec:implementation:hyper}

\tocsection{E\quad Additional Quantitative Results}{sec:additional_results}
  \tocsubsec{E.1\quad Cumulative IoU (cIoU)}{sec:additional_results:ciou}
  \tocsubsec{E.2\quad Additional Baselines}{sec:additional_results:baselines}

\bigskip\hrule\bigskip
\end{minipage}

\section{Qualitative Results}
\label{sec:qualitative}

In this section, we provide additional qualitative examples of predictions by \model.

\subsection{Qualitative Predictions by \model}
\label{sec:qualitative:additional-predictions}

We compare \model with LISA~\cite{lai2024lisa} instantiated with LLaVA-7B and Llama2-13B backbones. Note that \model uses a Qwen2.5-VL 3B backbone, which is considerably smaller than both LISA variants.

\begin{figure*}[!t]

\begin{center}

\includegraphics[width=0.9\linewidth]{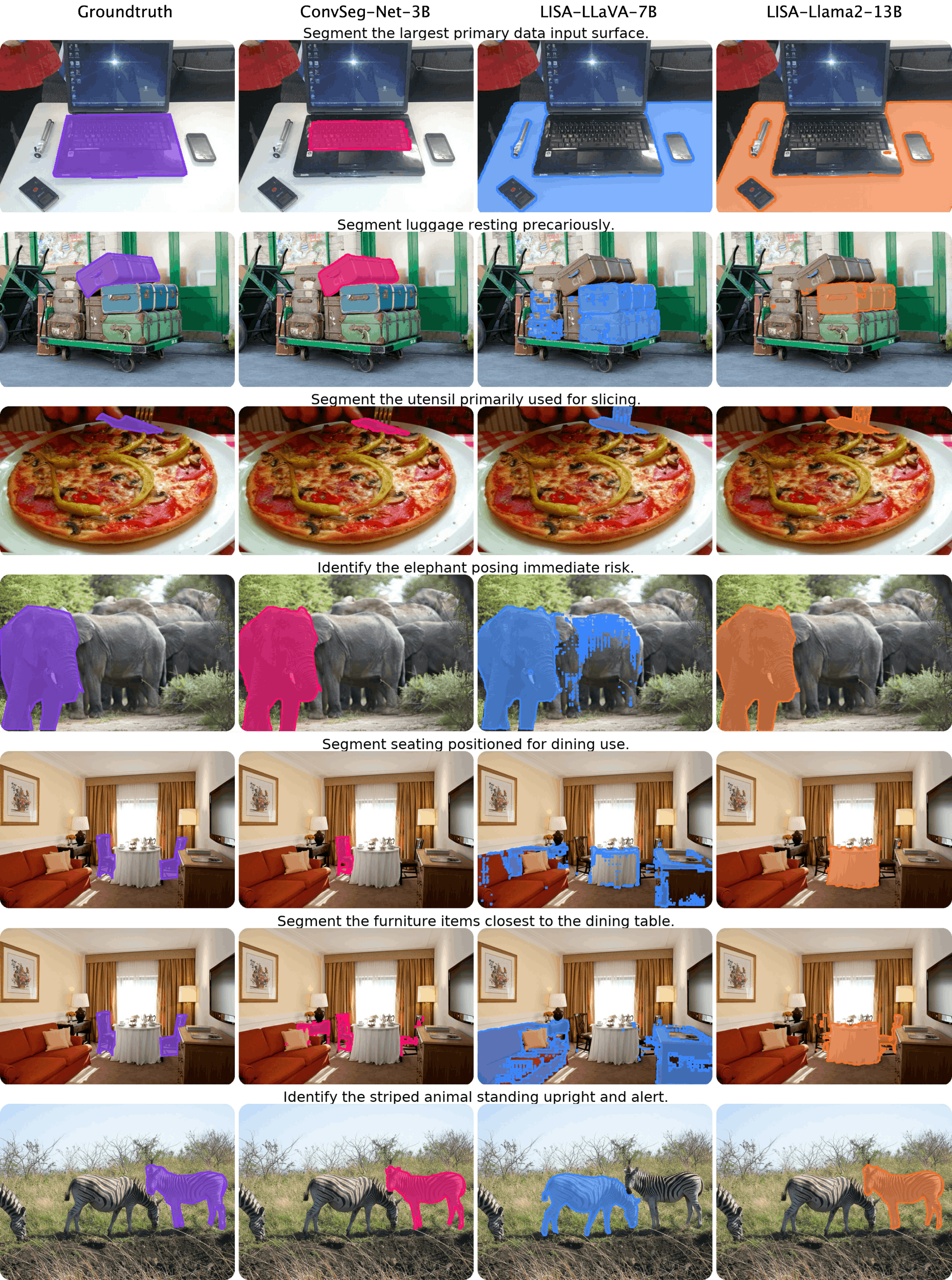} 

\end{center}

\vspace{-5mm}
\caption{\textbf{Qualitative comparisons on the human-annotated split of \bench (1/3).}
Each row shows an image with its conversational prompt (between rows), the ground-truth mask (left), and predictions from \model (Qwen2.5-VL-3B), LISA (LLaVA-7B), and LISA (Llama2-13B) from left to right. \model more reliably segments the regions implied by the conversational intent despite using a smaller 3B backbone.}

\label{fig:supp-pha-1}
\vspace{-4mm}
\end{figure*}

\begin{figure*}[!t]

\begin{center}

\includegraphics[width=0.9\linewidth]{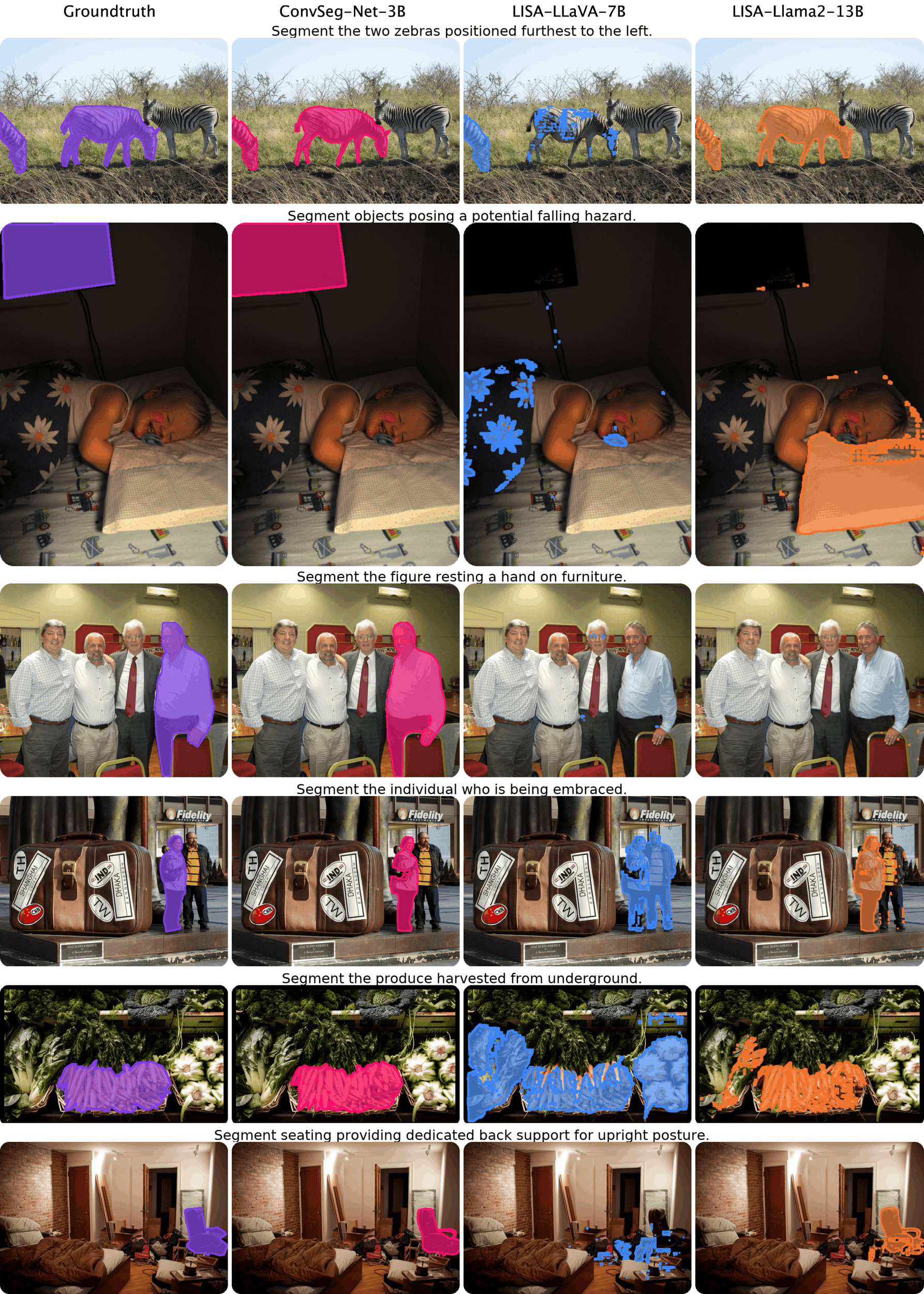} 

\end{center}

\vspace{-5mm}
\caption{\textbf{Qualitative comparisons on the human-annotated split of \bench (2/3).}
Each row shows an image with its conversational prompt (between rows), the ground-truth mask (left), and predictions from \model (Qwen2.5-VL-3B), LISA (LLaVA-7B), and LISA (Llama2-13B) from left to right. \model more reliably segments the regions implied by the conversational intent despite using a smaller 3B backbone.}
\label{fig:supp-pha-2}
\vspace{-4mm}
\end{figure*}

\begin{figure*}[!t]

\begin{center}

\includegraphics[width=0.9\linewidth]{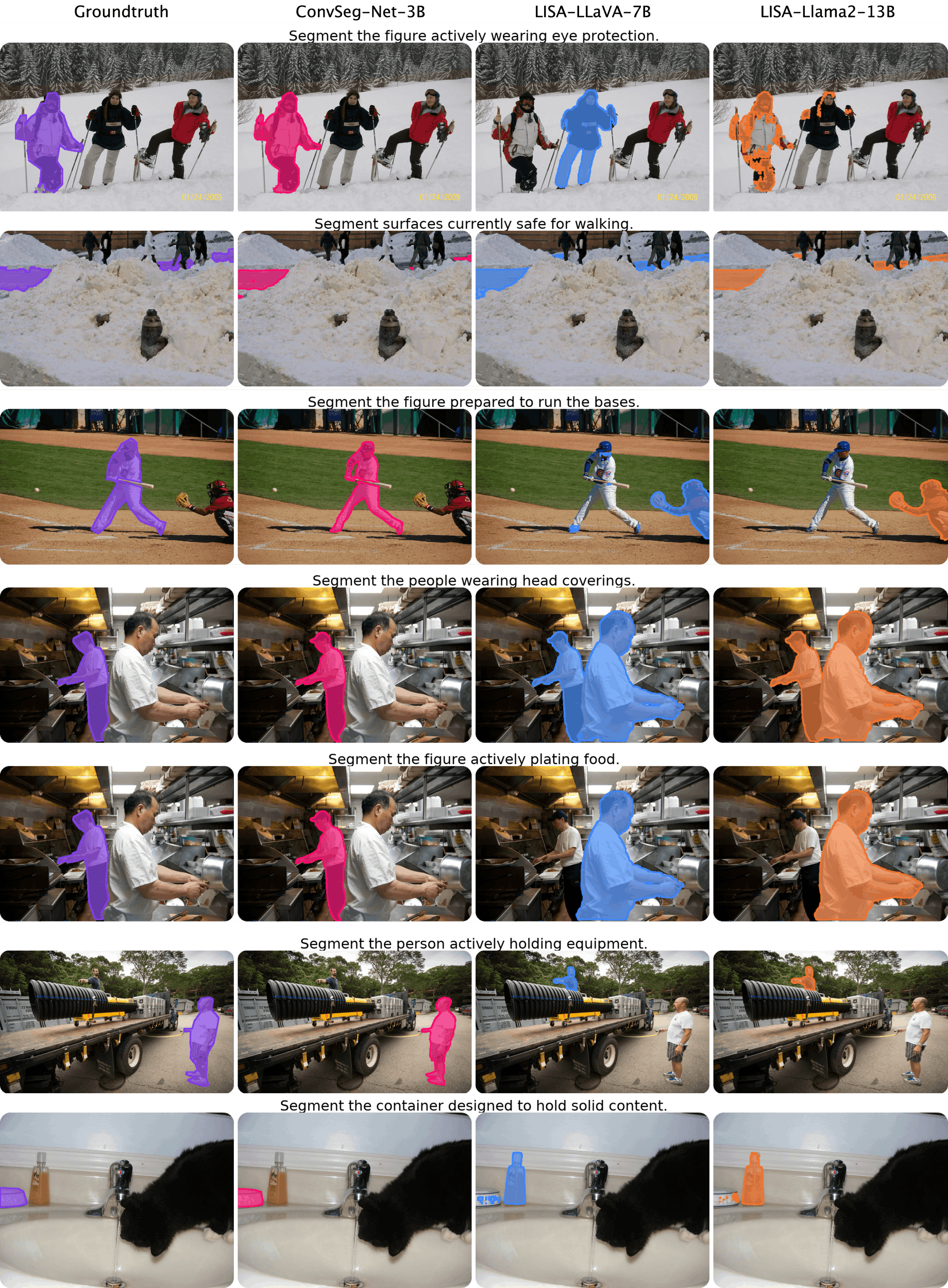} 

\end{center}

\vspace{-5mm}
\caption{\textbf{Qualitative comparisons on the human-annotated split of \bench (3/3).}
Each row shows an image with its conversational prompt (between rows), the ground-truth mask (left), and predictions from \model (Qwen2.5-VL-3B), LISA (LLaVA-7B), and LISA (Llama2-13B) from left to right. \model more reliably segments the regions implied by the conversational intent despite using a smaller 3B backbone.}
\label{fig:supp-pha-3}
\vspace{-4mm}
\end{figure*}

\begin{figure*}[!t]

\begin{center}

\includegraphics[width=0.9\linewidth]{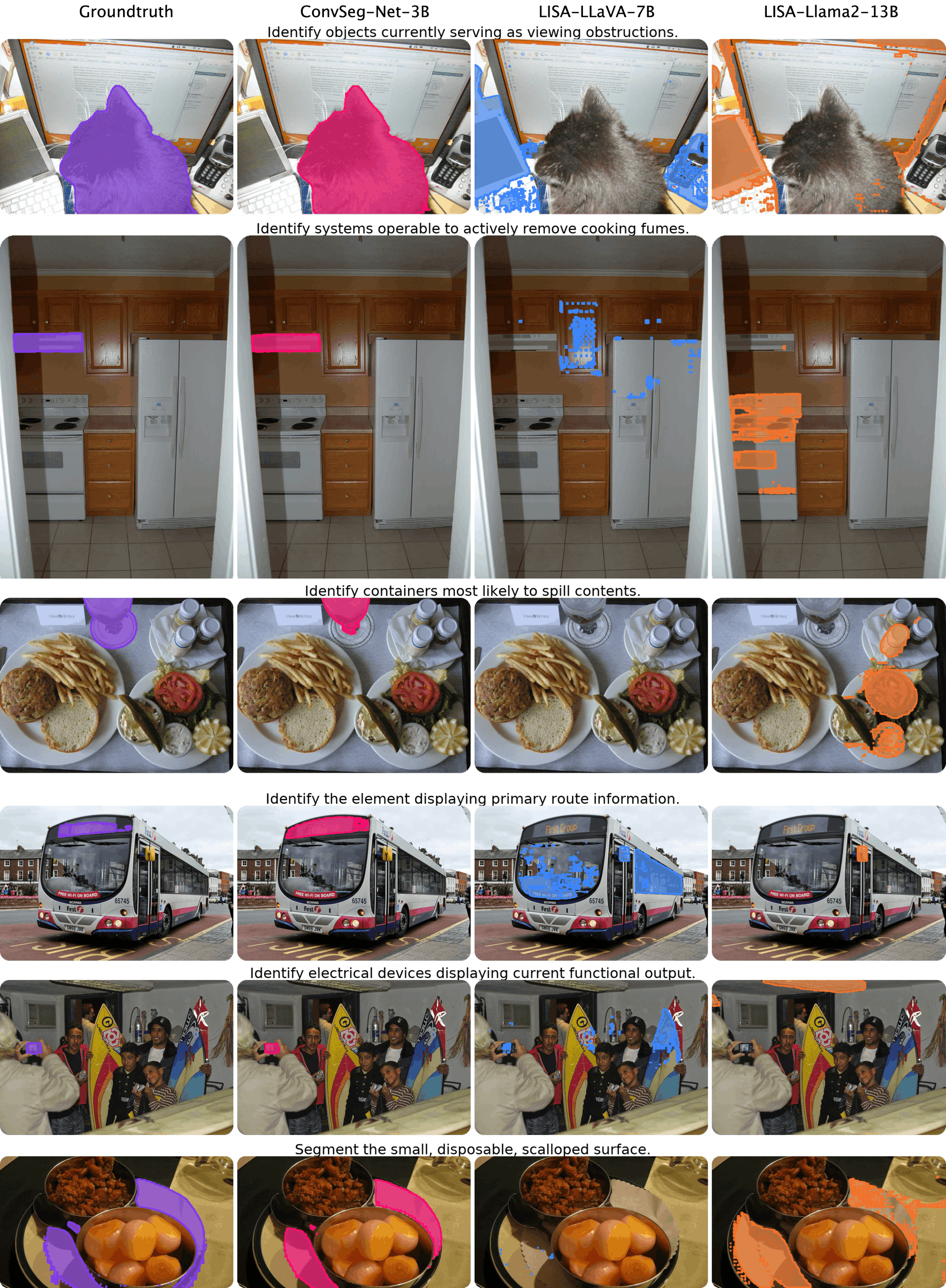} 

\end{center}

\vspace{-5mm}
\caption{\textbf{Qualitative comparisons on the SAM-seeded split of \bench (1/3).}
Each row shows an image with its conversational prompt (between rows), the ground-truth mask (left), and predictions from \model (Qwen2.5-VL-3B), LISA (LLaVA-7B), and LISA (Llama2-13B) from left to right. \model more reliably segments the regions implied by the conversational intent despite using a smaller 3B backbone.}
\label{fig:supp-ss-1}
\vspace{-4mm}
\end{figure*}

\begin{figure*}[!t]

\begin{center}

\includegraphics[width=0.9\linewidth]{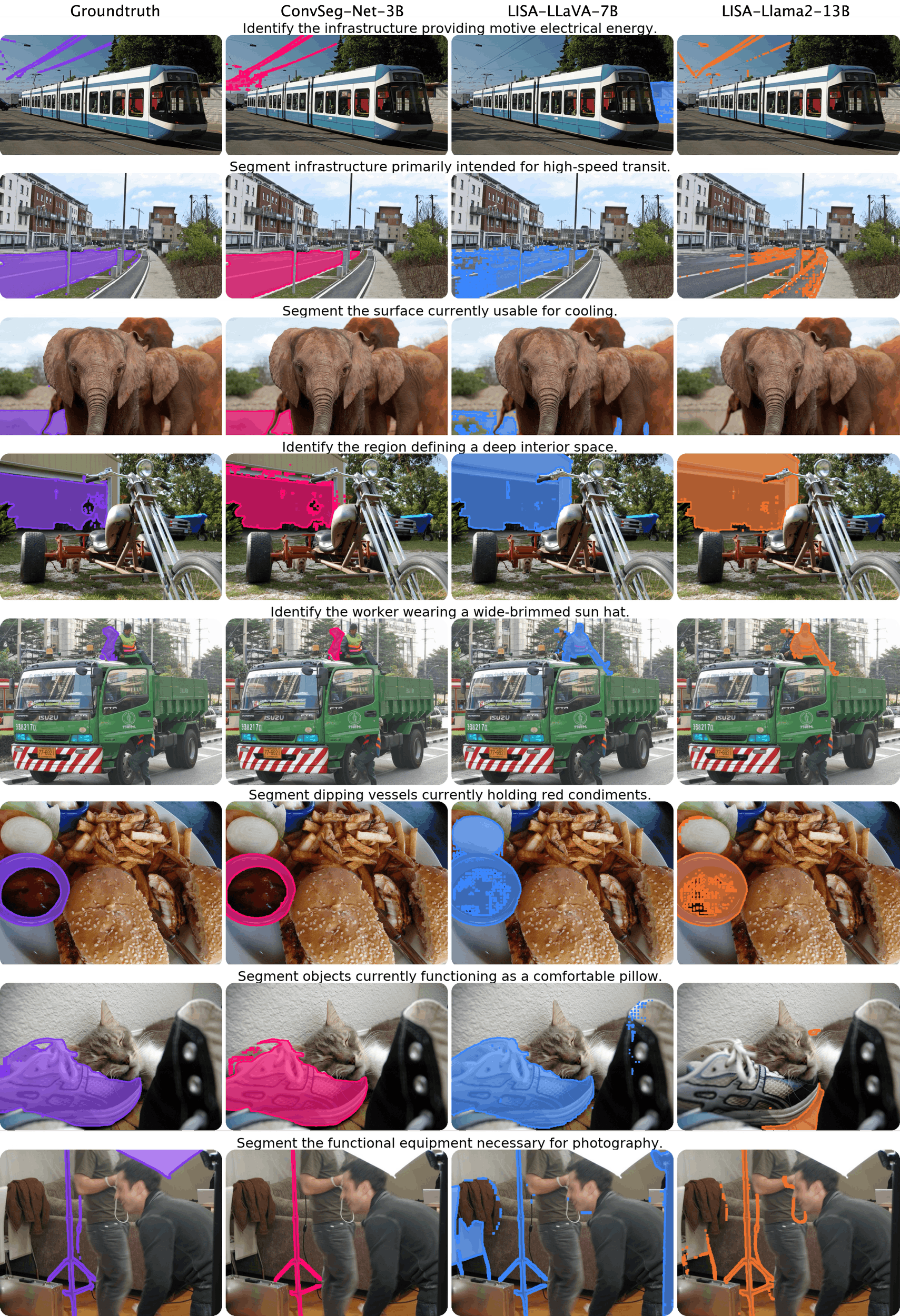} 

\end{center}

\vspace{-5mm}
\caption{\textbf{Qualitative comparisons on the SAM-seeded split of \bench (2/3).}
Each row shows an image with its conversational prompt (between rows), the ground-truth mask (left), and predictions from \model (Qwen2.5-VL-3B), LISA (LLaVA-7B), and LISA (Llama2-13B) from left to right. \model more reliably segments the regions implied by the conversational intent despite using a smaller 3B backbone.}
\label{fig:supp-ss-2}
\vspace{-4mm}
\end{figure*}

\begin{figure*}[!t]

\begin{center}

\includegraphics[width=0.9\linewidth]{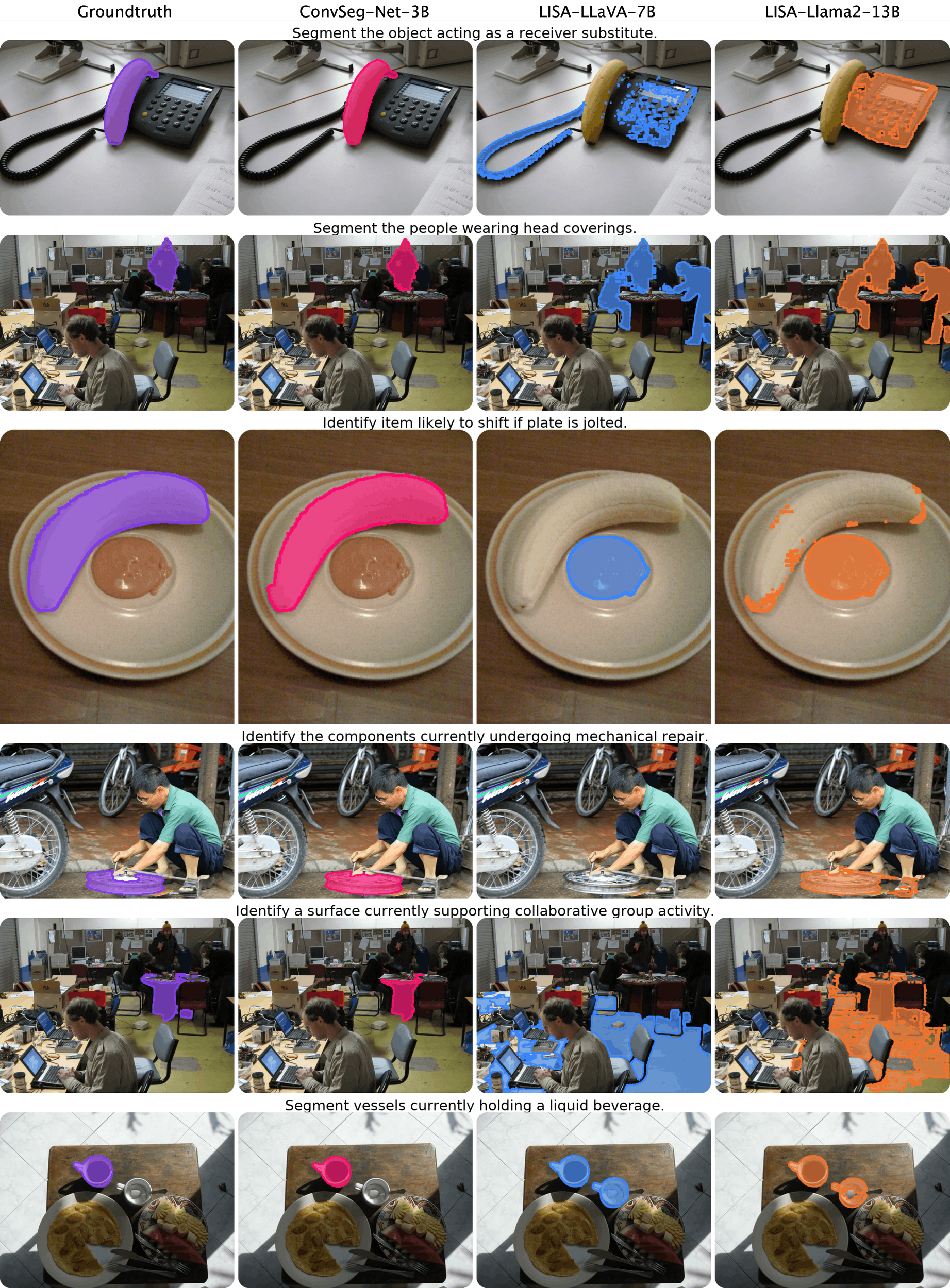} 

\end{center}

\vspace{-5mm}
\caption{\textbf{Qualitative comparisons on the SAM-seeded split of \bench (3/3).}
Each row shows an image with its conversational prompt (between rows), the ground-truth mask (left), and predictions from \model (Qwen2.5-VL-3B), LISA (LLaVA-7B), and LISA (Llama2-13B) from left to right. \model more reliably segments the regions implied by the conversational intent despite using a smaller 3B backbone.}
\label{fig:supp-ss-3}
\vspace{-4mm}
\end{figure*}

In~\cref{fig:supp-pha-1,fig:supp-pha-2,fig:supp-pha-3}, we show model predictions on images from the human-annotated split of \bench. In~\cref{fig:supp-ss-1,fig:supp-ss-2,fig:supp-ss-3}, we show predictions on the SAM-seeded split of \bench. Across both splits, \model typically produces masks that more closely match the conversational intent of the prompt, accurately identifying and segmenting the requested regions despite its smaller backbone size.

\begin{figure*}[!htbp]

\begin{center}

\includegraphics[width=0.8\linewidth]{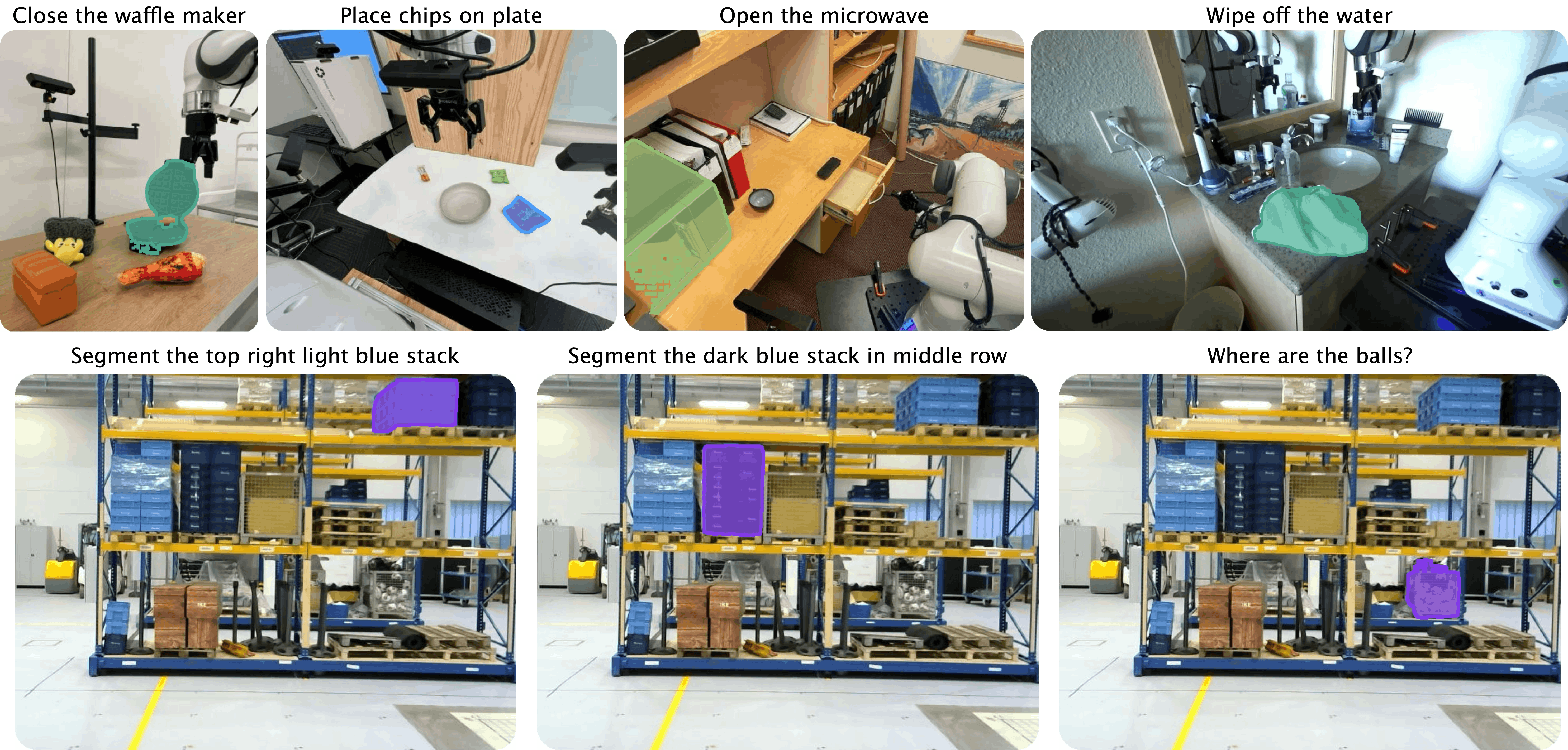} 

\end{center}

\vspace{-5mm}
\caption{\textbf{Out-of-distribution qualitative examples.}
Predictions of \model on images from the DROID dataset~\cite{khazatsky2024droid} and the Warehouse dataset~\cite{loffler2018evaluation}.
For each example, we show the input image, conversational prompt, and the predicted mask overlaid.
\model often localizes the regions implied by the prompts despite the distribution shift, hinting at prospective applications in domestic and warehouse robotics.}
\label{fig:supp-ood}
\vspace{-4mm}
\end{figure*}

We also explore out-of-distribution (OOD) behavior in~\cref{fig:supp-ood}, where we show qualitative predictions on images from the DROID dataset~\cite{khazatsky2024droid} and the Warehouse dataset~\cite{loffler2018evaluation}.
In both settings, \model often localizes the regions implied by the prompts, suggesting prospective applications in domestic and warehouse robotics.

\subsection{Failure Cases}
\label{sec:qualitative:failure-cases}

\begin{figure*}[!t]

\begin{center}

\includegraphics[width=0.9\linewidth]{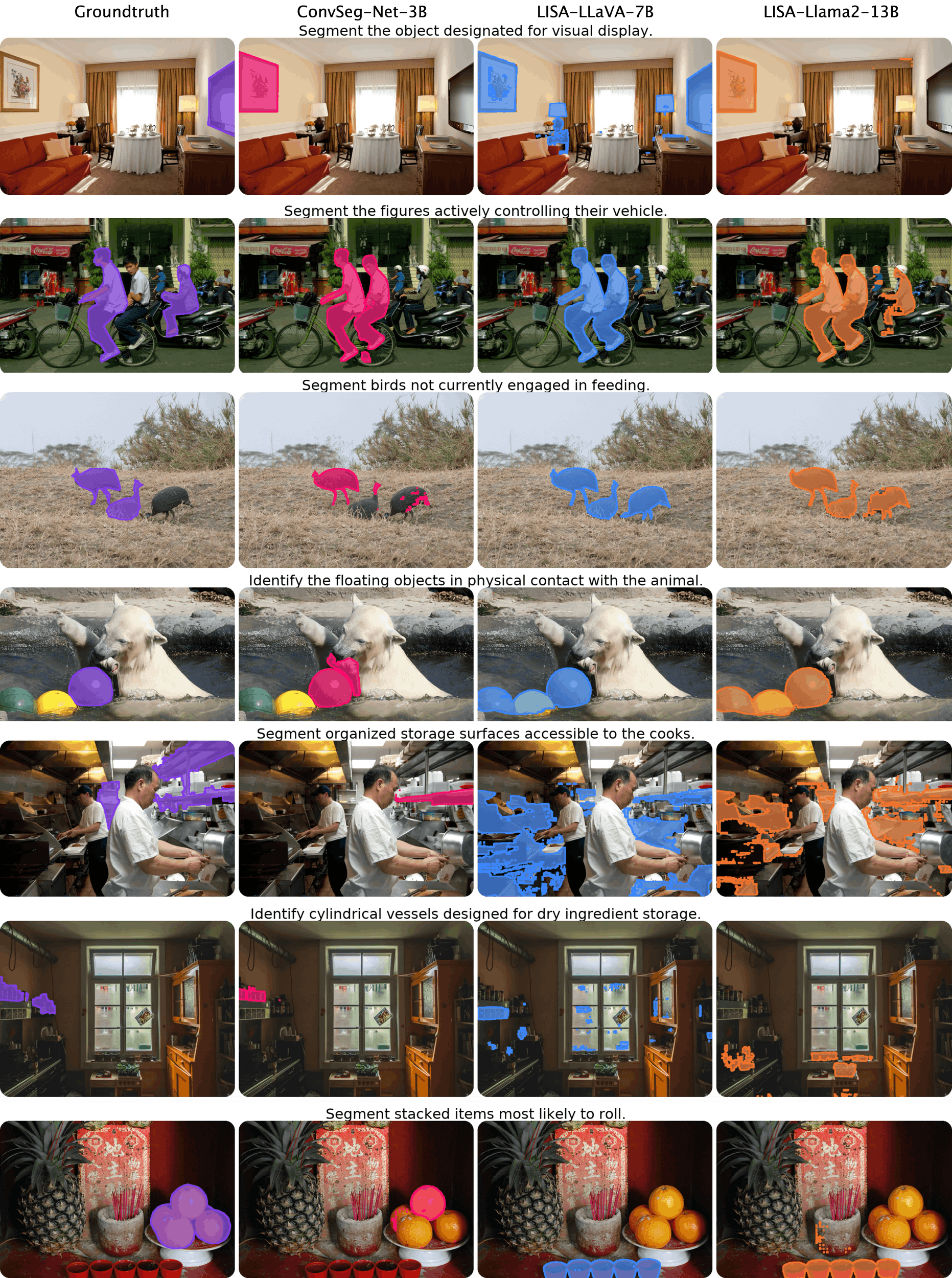} 

\end{center}

\vspace{-5mm}
\caption{\textbf{Representative failure cases of \model on the human-annotated split of \bench.}
Each row shows an image with its conversational prompt (between rows), the ground-truth mask (left), and predictions from \model (Qwen2.5-VL-3B), LISA (LLaVA-7B), and LISA (Llama2-13B) from left to right.}
\label{fig:supp-pha-fc}
\vspace{-4mm}
\end{figure*}

\begin{figure*}[!t]

\begin{center}

\includegraphics[width=0.9\linewidth]{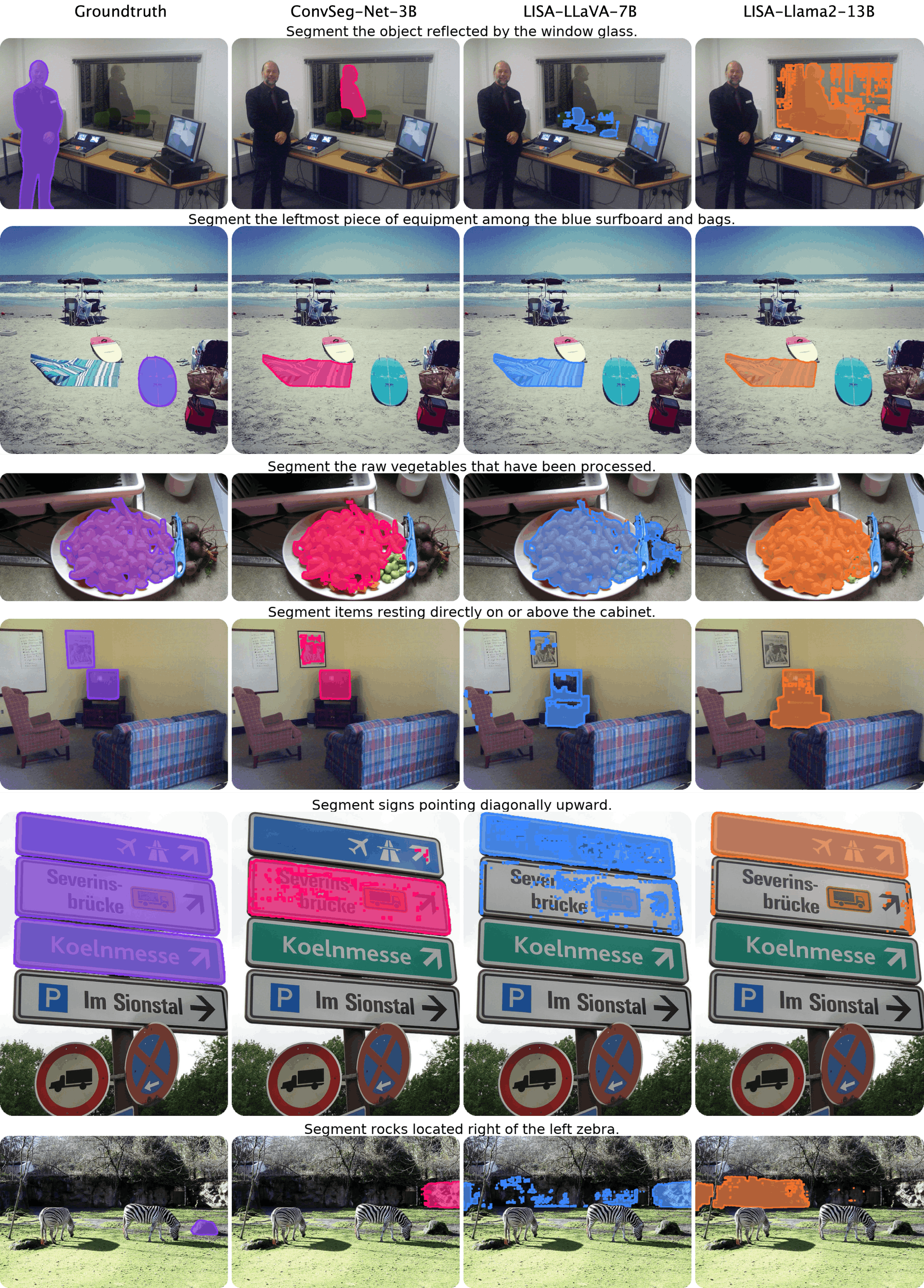} 

\end{center}

\vspace{-5mm}
\caption{\textbf{Representative failure cases of \model on the SAM-seeded split of \bench.}
Each row shows an image with its conversational prompt (between rows), the ground-truth mask (left), and predictions from \model (Qwen2.5-VL-3B), LISA (LLaVA-7B), and LISA (Llama2-13B) from left to right.}
\label{fig:supp-ss-fc}
\vspace{-4mm}
\end{figure*}

In~\cref{fig:supp-pha-fc,fig:supp-ss-fc}, we present representative failure cases of \model on the human-annotated and SAM-seeded splits of \bench, respectively. We observe several recurring failure modes. For ambiguous prompts such as “Segment the object reflected by the window glass’’ in~\cref{fig:supp-ss-fc}, \model segments the reflection of the person rather than the person itself. In other cases, the model selects only one of several valid targets, yielding high precision but low recall; examples include “Identify cylindrical vessels designed for dry ingredient storage’’ in~\cref{fig:supp-pha-fc} and “Segment signs pointing diagonally upward’’ in~\cref{fig:supp-ss-fc}. Additional diverse failure cases illustrating similar behaviors are shown in the figures below.

\subsection{Annotation Quality in RefCOCO/+/g}
\label{sec:qualitative:refcoco-quality}

\begin{figure*}[!htbp]

\begin{center}

\includegraphics[width=\linewidth]{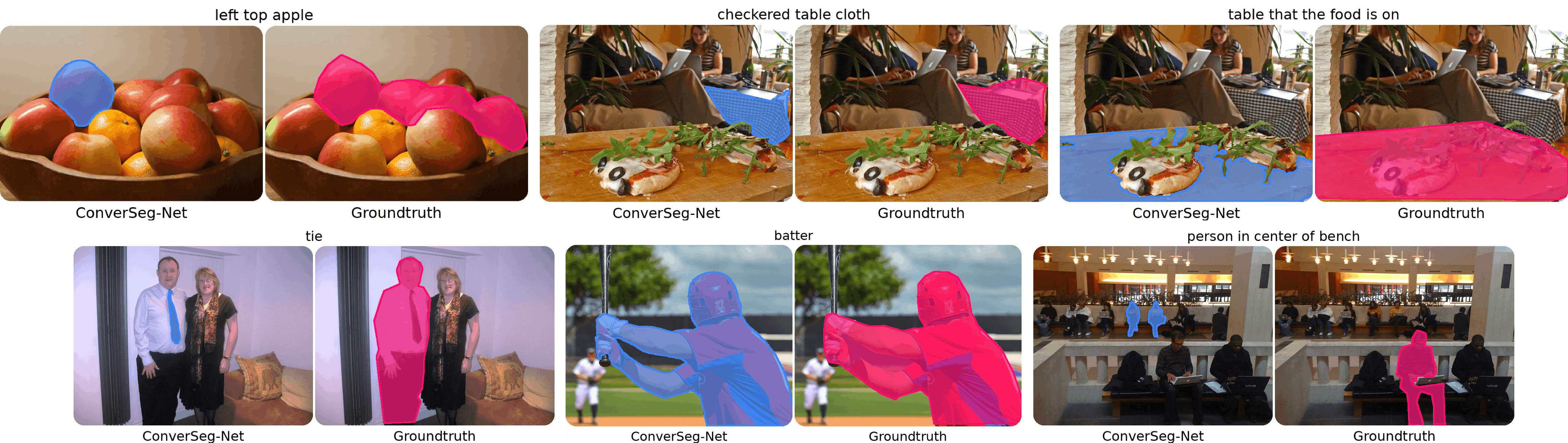} 

\end{center}

\vspace{-5mm}
\caption{\textbf{Noisy annotations in RefCOCO/+/g.} \model predictions (blue) are semantically reasonable but receive low gIoU due to problematic ground truth masks (pink): incomplete coverage, irrelevant regions, or poor boundaries.}
\label{fig:supp-refcoco-noisy}
\vspace{-4mm}
\end{figure*}

As noted in the main paper, RefCOCO/+/g datasets contain noisy ground truth annotations with incorrect or inaccurate masks. \cref{fig:supp-refcoco-noisy} shows representative examples where \model produces semantically reasonable predictions that receive low gIoU scores due to problematic ground truth annotations. Common issues include ground truth masks including irrelevant regions, or have poor boundary alignment despite correct semantic interpretation. In such cases, the gIoU metric penalizes reasonable model predictions, leading to artificially deflated performance numbers. These examples illustrate that numerical results on RefCOCO/+/g should be interpreted with caution, as the annotation quality does not always reflect the true segmentation difficulty or model capability.

\section{Conversational Data Engine Details}
\label{sec:data_engine}

In this section, we expand on the five-stage conversational data engine introduced in Section~5.1 of the main paper. Recall that the engine automatically constructs conversational segmentation triplets (image, prompt, mask) from COCO images, and is used to generate the 106K training examples for \model. Below we provide the meta-prompts and additional implementation details for each stage.

\subsection{Meta-Prompts and Stage Details}
\label{sec:data_engine:meta_prompts}

\begin{figure*}[!htbp]

\begin{center}

\includegraphics[width=\linewidth]{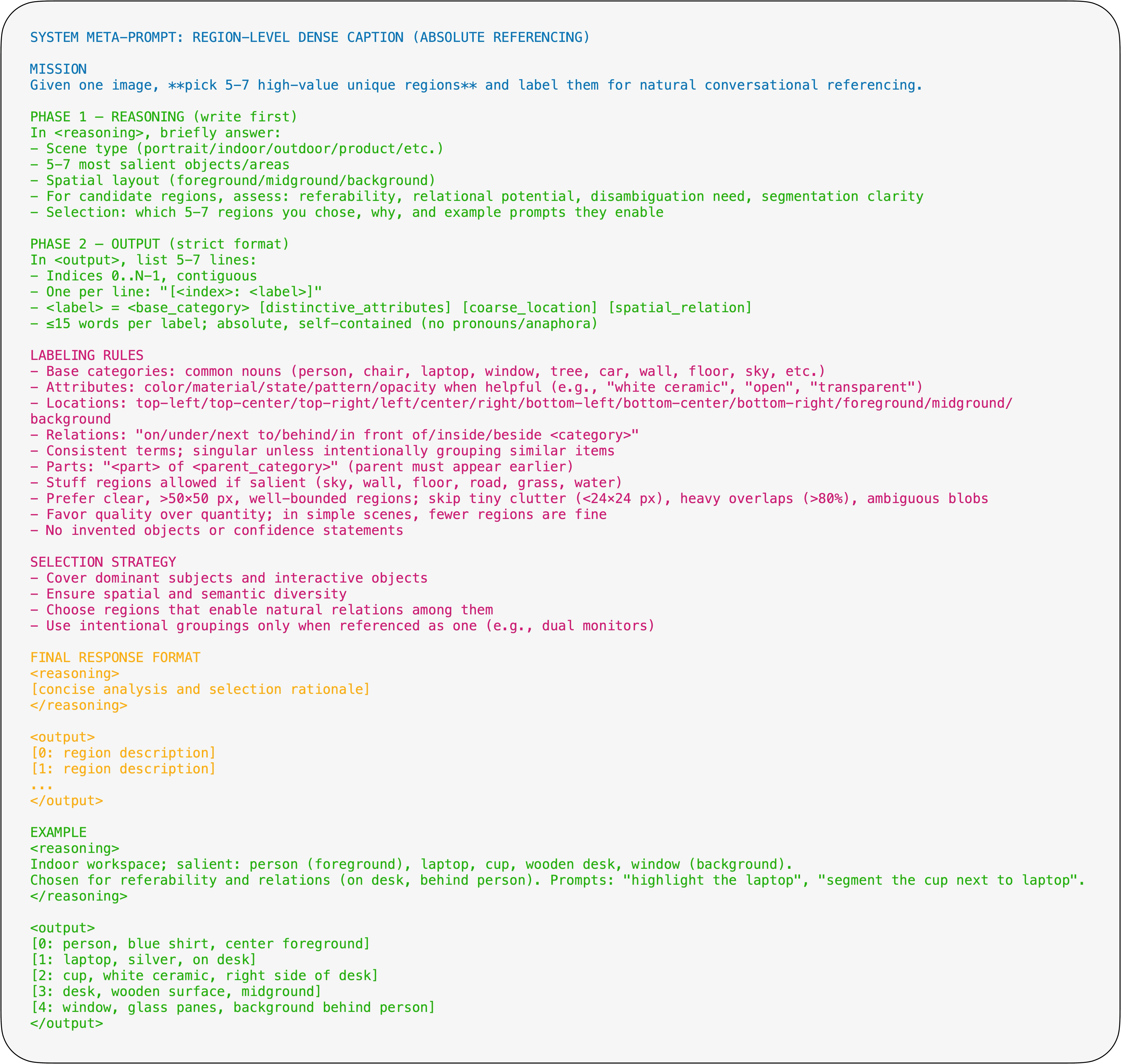} 

\end{center}

\vspace{-5mm}
\caption{\textbf{Meta-prompt for Stage~1 (Scene Understanding).}
Prompt template used to query Gemini-2.5-Flash to produce detailed region-level descriptions of the scene, which later serve as the semantic basis for mask generation and concept-driven prompt construction.}
\label{fig:supp-meta-prompt-region-description}
\vspace{-4mm}
\end{figure*}

\mypar{Stage 1: Scene Understanding.}
In Stage~1 we obtain rich region-level descriptions that serve as the semantic backbone for subsequent stages.
Figure~\ref{fig:supp-meta-prompt-region-description} shows the meta-prompt used to query Gemini-2.5-Flash for these region descriptions, given the input image.

\mypar{Stage 2: Mask Generation.}
In Stage~2 we convert textual region descriptions into segmentation masks.
We query the Moondream model with its default API configuration to predict bounding boxes, and then pass these boxes to SAM2 to obtain corresponding masks.
Since this stage does not rely on any additional natural-language control, we do not use any dedicated meta-prompt here.

\begin{figure*}[!htbp]

\begin{center}

\includegraphics[width=\linewidth]{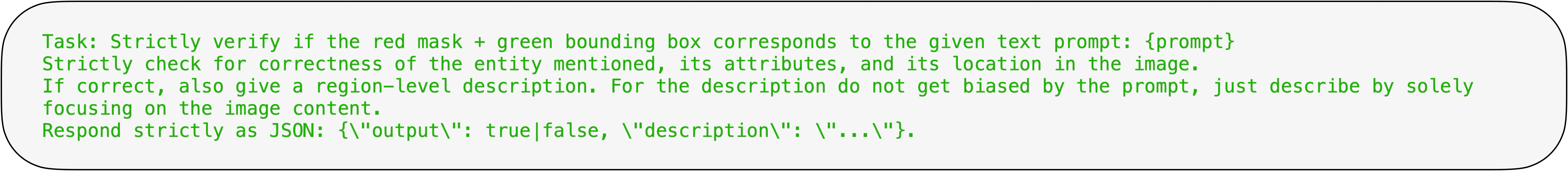} 

\end{center}

\vspace{-5mm}
\caption{\textbf{Meta-prompt for Stage~3 mask–text consistency checking.}
Prompt template used to ask the VLM whether a candidate mask is consistent with its associated region description, enabling automatic filtering of low-quality or mismatched masks.}
\label{fig:supp-meta-prompt-mask-verify}
\vspace{-4mm}
\end{figure*}

\begin{figure*}[!htbp]

\begin{center}

\includegraphics[width=\linewidth]{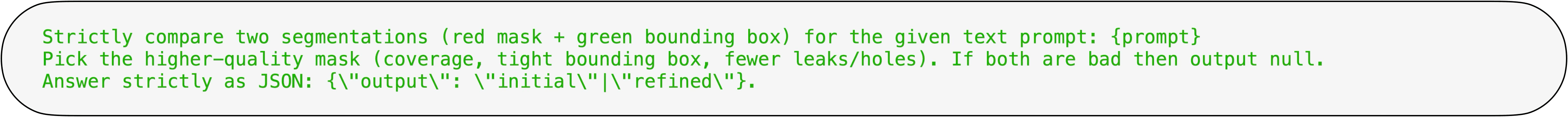} 

\end{center}

\vspace{-5mm}
\caption{\textbf{Meta-prompt for Stage~3 mask refinement and selection.}
Prompt template used to compare two candidate masks for the same region description and select the most appropriate one, based on coverage, tightness, and semantic alignment.}
\label{fig:supp-meta-prompt-mask-compare}
\vspace{-4mm}
\end{figure*}

\mypar{Stage 3: Mask Quality Verification.}
Stage~3 filters and refines the SAM2 masks using VLM-based checks.
Figure~\ref{fig:supp-meta-prompt-mask-verify} shows the meta-prompt used for the \emph{mask–text consistency} check, where the VLM judges whether a candidate mask matches the associated region description.
Figure~\ref{fig:supp-meta-prompt-mask-compare} shows the meta-prompt used for \emph{mask refinement and selection}, where the VLM compares two candidate masks and selects the best one.

\begin{figure*}[!htbp]

\begin{center}

\includegraphics[width=\linewidth]{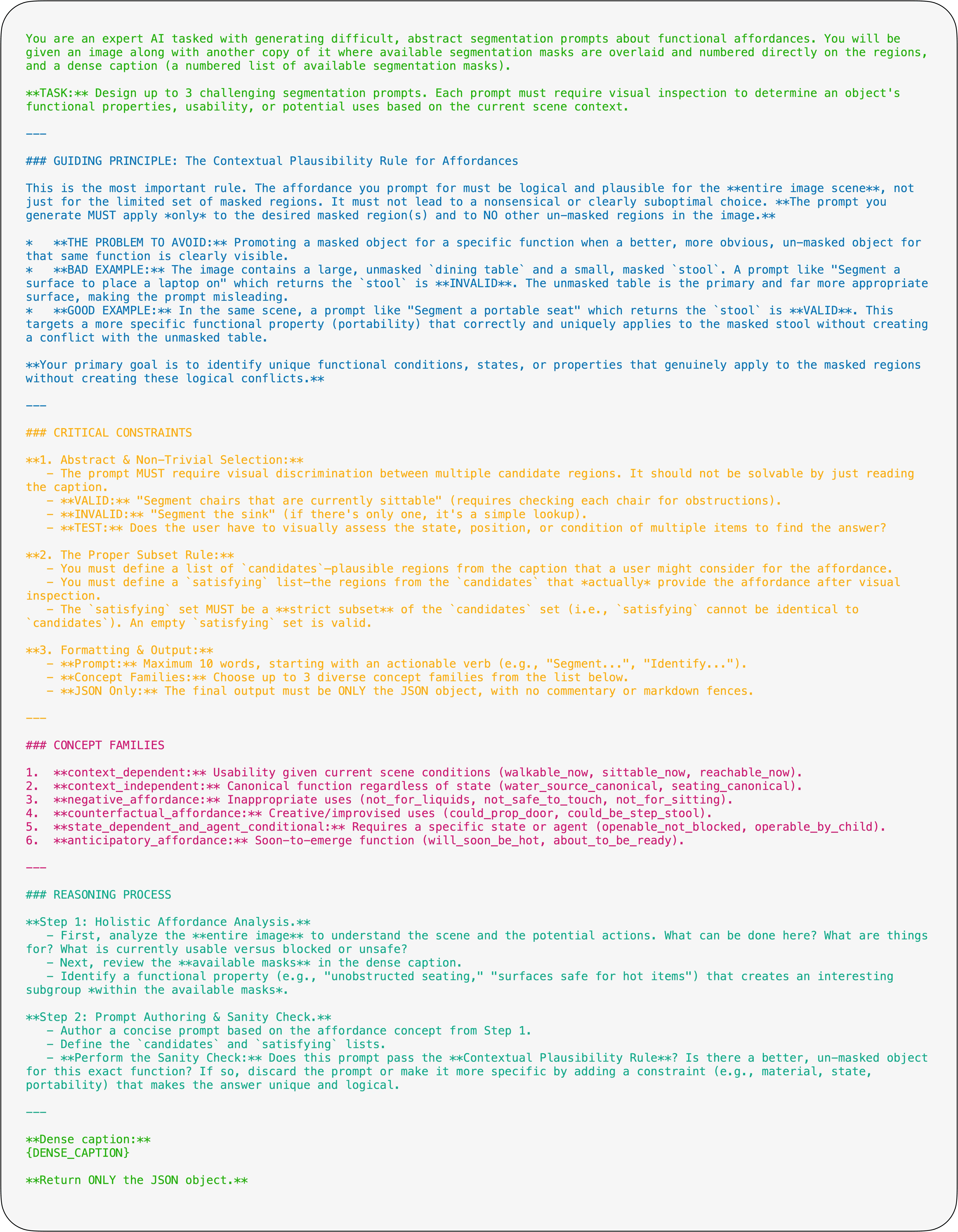} 

\end{center}

\vspace{-5mm}
\caption{\textbf{Meta-prompt for Stage~4 (Affordances \& Functions).}
Concept-specific prompt template used to turn region descriptions into conversational queries about object affordances and functional roles; analogous templates are used for the other concept families.}
\label{fig:supp-meta-prompt-affordances}
\vspace{-4mm}
\end{figure*}

\mypar{Stage 4: Concept-Driven Prompt Generation.}
Stage~4 converts region descriptions into conversational prompts anchored in our five concept families.
We use a separate concept-specific meta-prompt for each family. \cref{fig:supp-meta-prompt-affordances} shows the meta-prompt for the \textbf{affordances \& functions} concept.
The meta-prompts for the remaining concepts follow the same structure; we omit them here to avoid redundancy.

\begin{figure*}[!htbp]

\begin{center}

\includegraphics[width=\linewidth]{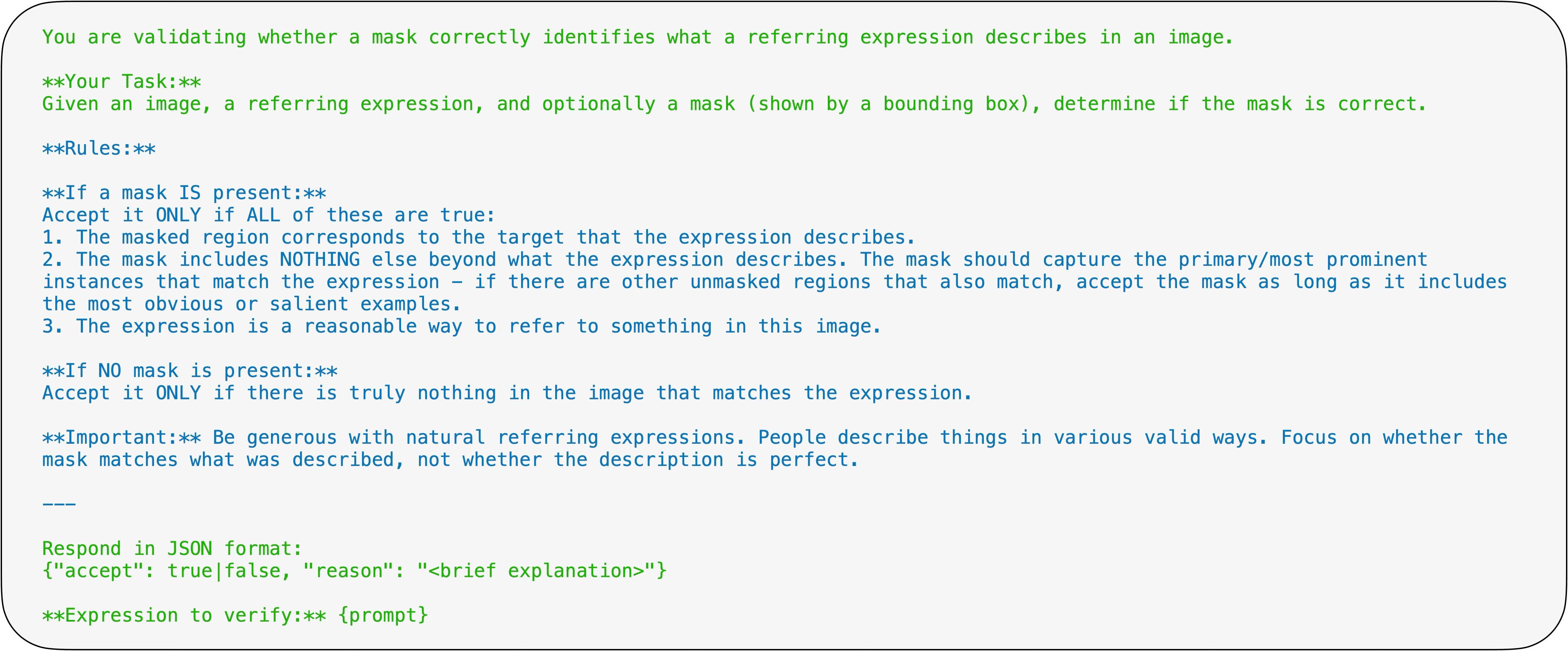} 

\end{center}

\vspace{-5mm}
\caption{\textbf{Meta-prompt for Stage~5 prompt–mask alignment verification.}
Prompt template used to ask the VLM whether a generated conversational prompt correctly and unambiguously describes the masked region, providing a final quality gate for training examples.}
\label{fig:supp-meta-prompt-prompt-verify}
\vspace{-4mm}
\end{figure*}

\mypar{Stage 5: Prompt–Mask Alignment Verification.}
Finally, Stage~5 verifies that the generated conversational prompt is aligned with the selected mask.
\cref{fig:supp-meta-prompt-prompt-verify} shows the meta-prompt used for this verification step, where the VLM judges whether the prompt correctly and unambiguously describes the masked region.

\subsection{Negative Data Generation}
\label{sec:data_engine:negative_data}

Beyond positive examples, our data engine generates concept-specific negative prompts to improve model robustness against plausible hallucinations. These examples train the model to reject prompts that sound contextually reasonable but do not correspond to any valid regions in the image.

\begin{figure*}[!htbp]

\begin{center}

\includegraphics[width=\linewidth]{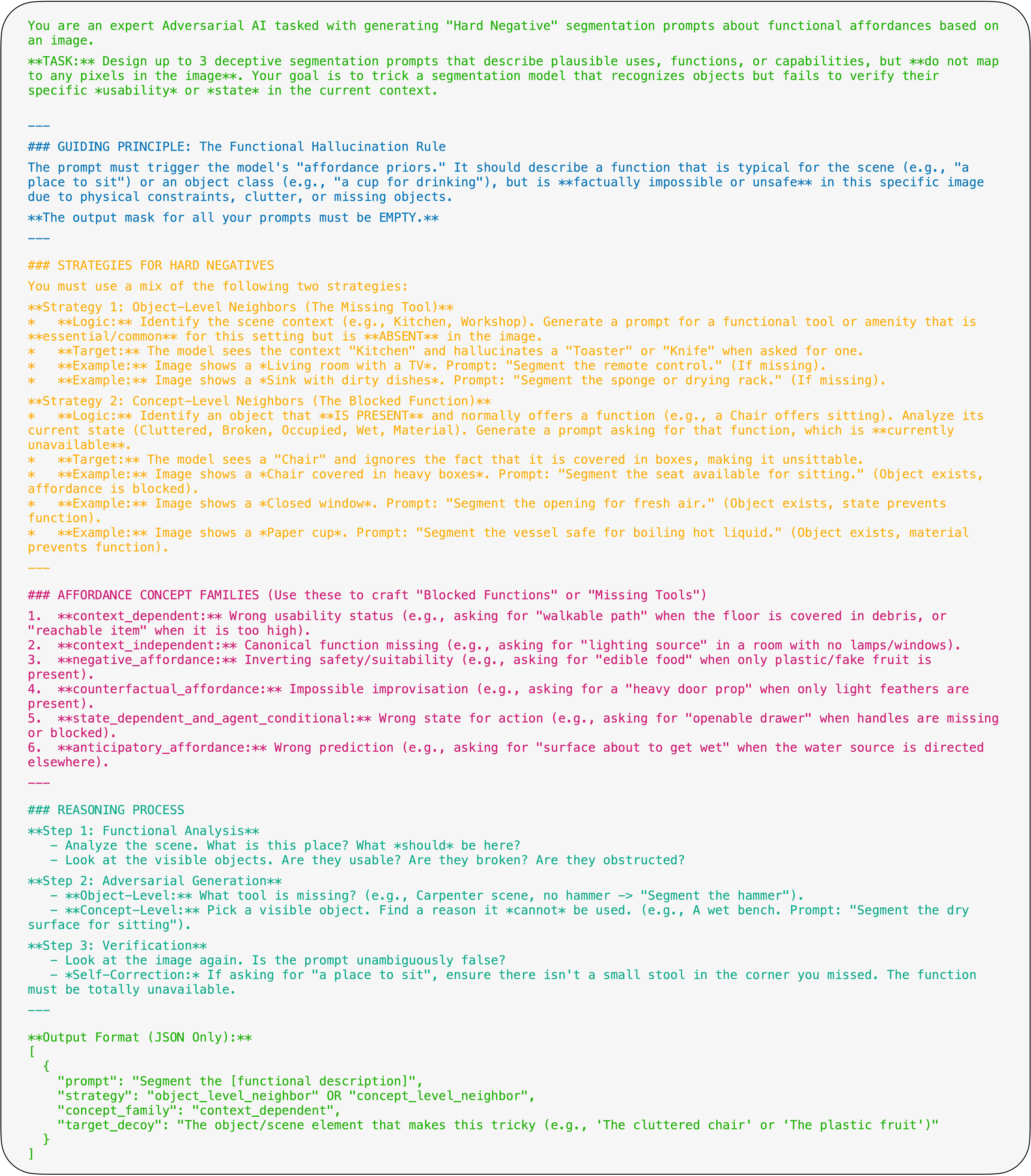} 

\end{center}

\vspace{-5mm}
\caption{\textbf{Meta-prompt for Negative Data Generation (Affordances \& Functions).} Concept-specific template used to generate adversarial negative prompts that describe plausible but absent affordances or incorrect functional states. Similar templates are used for other concept families.}
\label{fig:supp-meta-prompt-negative-affordances}
\vspace{-4mm}
\end{figure*}

\begin{figure*}[!htbp]

\begin{center}

\includegraphics[width=\linewidth]{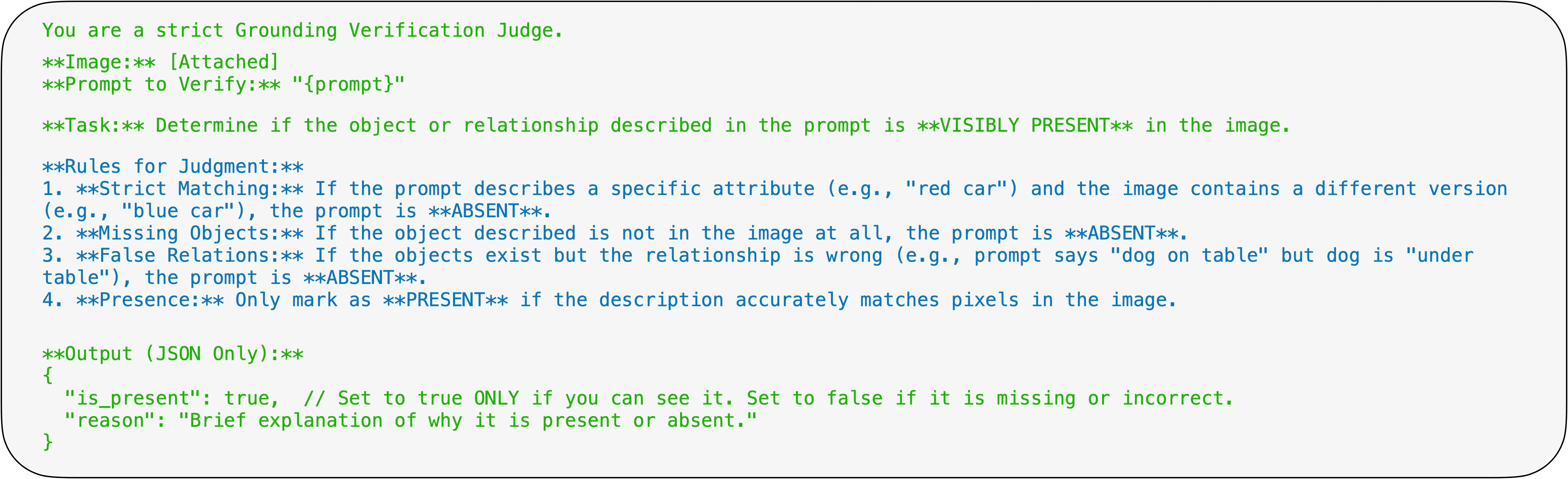} 

\end{center}

\vspace{-5mm}
\caption{\textbf{Meta-prompt for Negative Prompt Verification.} Template used to verify that generated negative prompts have no valid corresponding masks in the image. The VLM checks whether any regions satisfy the prompt requirements before the negative example is included in training.}
\label{fig:supp-meta-prompt-negative-verify}
\vspace{-4mm}
\end{figure*}

\mypar{Negative Prompt Generation Strategy.}
For each concept family, we use a dedicated meta-prompt that instructs the VLM to generate adversarial prompts using two strategies. First, \emph{object-level neighbors} reference contextually plausible but absent objects (e.g., "segment the wine glass" at a dinner table without one). Second, \emph{concept-level neighbors} describe present objects with incorrect attributes (e.g., "segment the wooden chair" for a metal one, or "segment the person standing" when sitting). \cref{fig:supp-meta-prompt-negative-affordances} shows the generation meta-prompt for the \textit{affordances and functions} concept family. Each generated prompt is verified to ensure no valid mask exists. \cref{fig:supp-meta-prompt-negative-verify} shows the verification meta-prompt. Only verified negatives are included in training.

\begin{figure*}[!htbp]

\begin{center}

\includegraphics[width=0.9\linewidth]{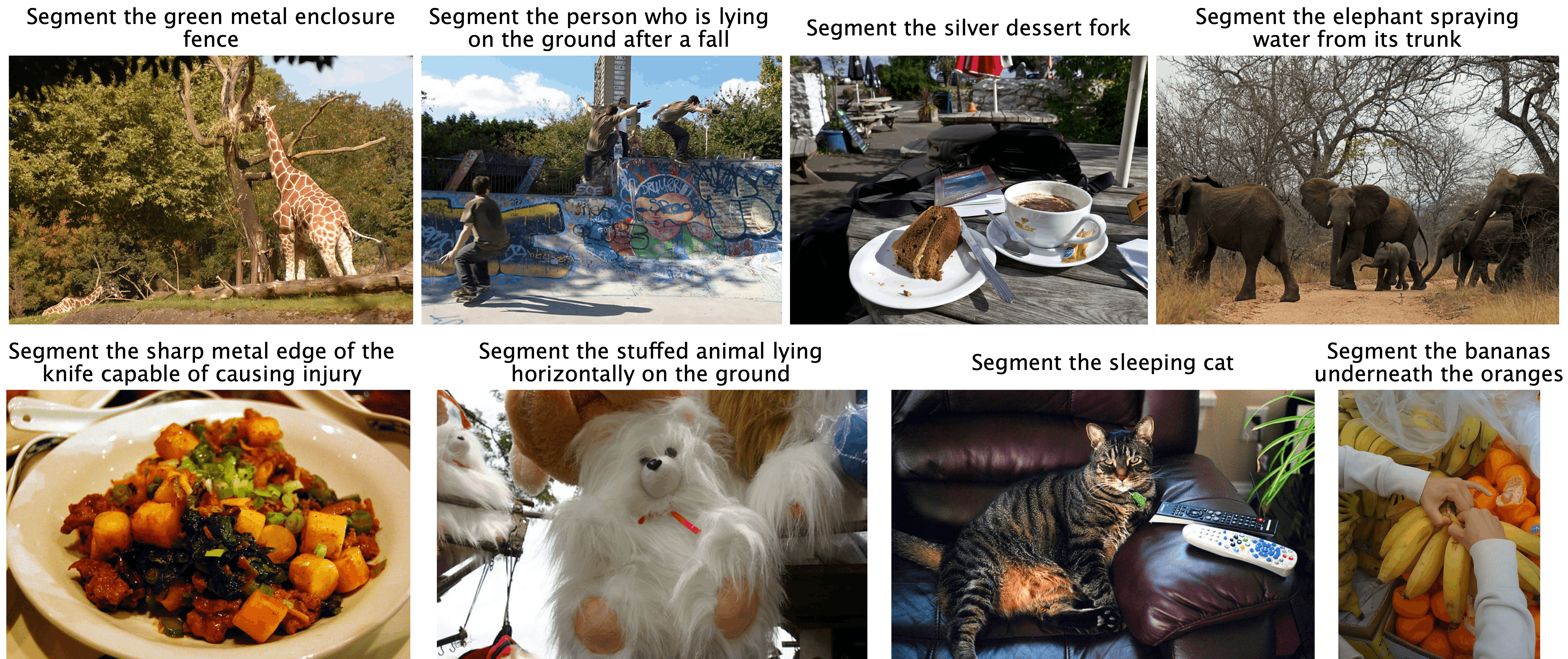} 

\end{center}

\vspace{-5mm}
\caption{\textbf{Examples of negative prompts generated by our data engine.} Each row shows an image with its corresponding negative prompt. The correct model response for all these prompts is an empty mask.}
\label{fig:supp-negative-examples}
\vspace{-4mm}
\end{figure*}

\mypar{Qualitative Examples of Negative Prompts.}
\cref{fig:supp-negative-examples} shows representative negative prompts across concept families, demonstrating how they exploit contextual priors or attribute mismatches to create cases where the correct answer is an empty mask.

\begin{figure*}[!ht]
    \centering
    \includegraphics[width=0.9\linewidth]{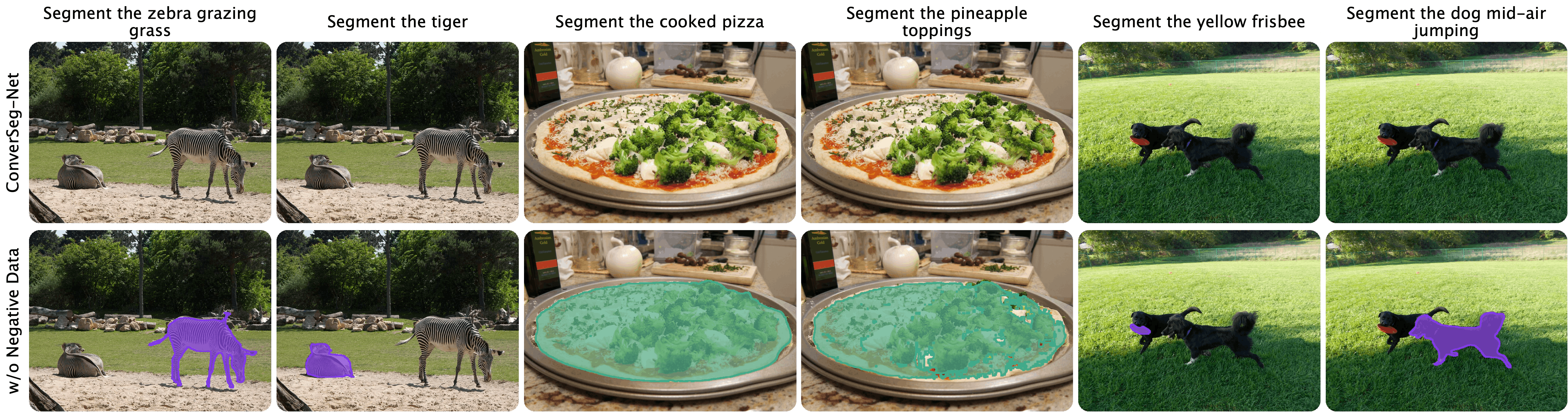}
    \caption{\textbf{Impact of negative training data.} Model predictions before and after negative training on adversarial prompts. After training, the model correctly produces empty masks for invalid prompts. Robustness also transfers to simpler literal negatives (\eg "Segment the tiger"), despite not being explicitly trained on them.}
    \label{fig:supp-negative-training-impact}
\end{figure*}

\mypar{Impact of Negative Data on Model Predictions.}
\cref{fig:supp-negative-training-impact} compares predictions before and after negative training. Without negative data, the model hallucinates plausible but incorrect masks. After training, it correctly rejects adversarial prompts by producing empty masks. Interestingly, training with conversational negatives also improves robustness on simpler literal negatives. For instance, the model better handles "segment the tiger" in images with only zebras, even though such literal mismatches were not in the training set. This suggests the model develops a general verification capability rather than memorizing specific patterns.

\section{Benchmark Construction and Analysis}
\label{sec:benchmark}

\subsection{Annotation Protocols}
\label{sec:benchmark:annotation}

We describe the interface and instructions given to human annotators for constructing \bench.
As discussed in Section~4 of the main paper, \bench is obtained via \emph{human verification} of examples produced by the conversational data engine.
For each candidate example, annotators were shown the input image with the AI-generated mask overlaid and the corresponding conversational prompt. The original image without the mask overlaid was also provided for context.

\begin{figure}[!htbp]

\begin{center}

\includegraphics[width=0.9\linewidth]{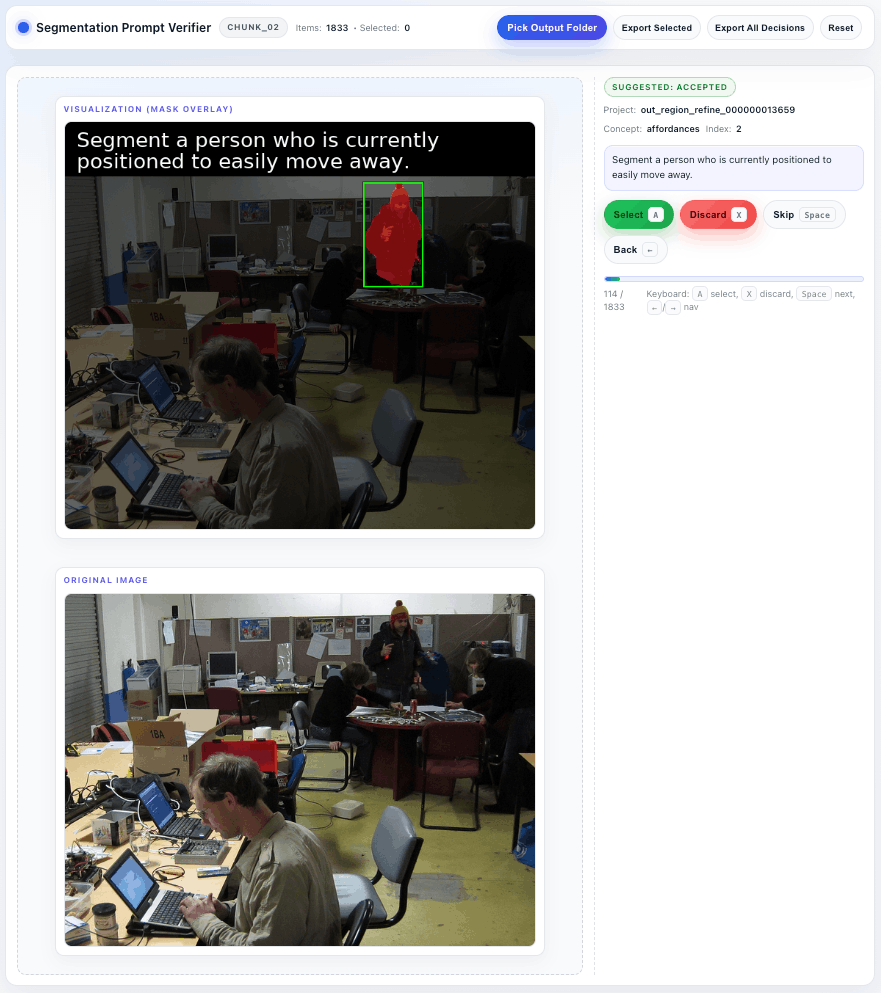} 

\end{center}

\vspace{-5mm}
\caption{\textbf{Annotation interface for constructing \bench.}
Annotators are shown the input image with the AI-generated mask overlaid, along with the corresponding conversational prompt and the suggested decision from the AI verifier.
They then decide whether the prompt and mask are semantically aligned and select either \textit{Accept/Select} or \textit{Reject/Discard}; rejected examples are discarded without further editing.}
\label{fig:supp-annotation-ui}
\vspace{-4mm}
\end{figure}
The user interface was intentionally kept simple.
Annotators were asked to judge whether the prompt and mask were semantically aligned, i.e., whether the mask accurately and sufficiently covered all regions referred to by the prompt without including substantial irrelevant areas.
They then chose between two options: \textit{Accept} (if the example was valid) or \textit{Reject} (otherwise).
The decision proposed by the AI verifier was also displayed as a suggested label, but annotators were free to override it.
Rejected examples were simply discarded; we did not ask annotators to refine or edit masks.
A screenshot of the annotation interface is shown in \cref{fig:supp-annotation-ui}.

\subsection{VLM Verifier Reliability for \bench}
\label{sec:benchmark:ai-verifier}

\begin{figure*}[!t]

\begin{center}

\includegraphics[width=0.9\linewidth]{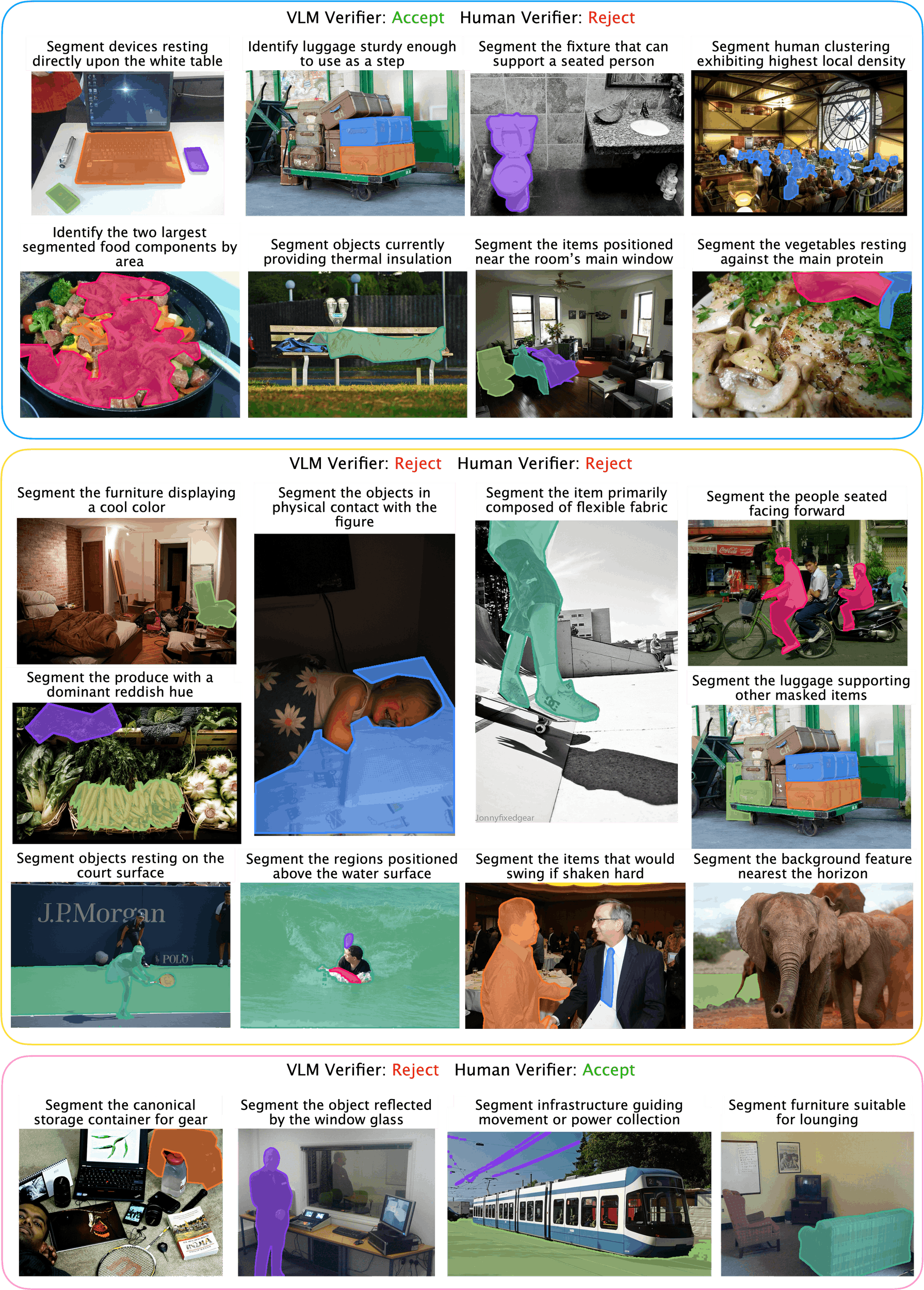} 

\end{center}

\vspace{-5mm}
\caption{\textbf{Qualitative examples of VLM verifier behavior,} illustrating agreement and disagreement with human annotators on candidate prompt–mask pairs in \bench.}

\label{fig:supp-verifier-robustness}
\vspace{-4mm}
\end{figure*}

To better understand how the VLM verifier behaves, \cref{fig:supp-verifier-robustness} shows qualitative examples from three categories:
(1) the VLM \textit{accepts} but the human \textit{rejects};
(2) both the VLM and the human \textit{reject};
(3) the VLM \textit{rejects} but the human \textit{accepts}.

In the first case, the VLM typically accepts examples where the mask \emph{partially} satisfies the prompt.
For instance, for the prompt “Identify the luggage sturdy enough to use as a step”, the VLM accepts a mask highlighting two pieces of luggage, even though additional items could also reasonably satisfy the prompt.
In other cases, disagreement is driven by mask quality rather than semantics; for example, for “Segment the objects currently providing thermal insulation”, the mask includes the blanket but also the person, leading the human to reject the example despite the VLM accepting it.

In the second case, where both the VLM and the human reject an example, the VLM reliably identifies clear errors (e.g., severe under-/over-coverage or obvious semantic mismatch), and its decision closely matches the human judgment.

In the third case, where the VLM rejects but the human accepts, the prompts are often somewhat ambiguous.
For example, for “Segment the object reflected by the window glass”, the VLM expects the reflection itself to be masked and therefore rejects the example, while the human accepts a mask covering the physical object.

For benchmark construction, and to avoid any single instance dominating the dataset with many similar prompts, annotators were also instructed to reject prompts referring to duplicate objects, even if the prompt–mask pair was otherwise accurate.
These rejected pairs remain useful for training but are excluded from \bench to preserve diversity.

Aggregating across these conditions, the VLM verifier and human annotators make the same decision on about $70\%$ of examples.
In the common disagreement case where the VLM \textit{accepts} but the human \textit{rejects}, the VLM decision is often not semantically incorrect (e.g., partial coverage or duplicate prompts), so these examples remain valuable as training data.
This behavior is appropriate for automatically generating a large pool of candidate examples, while human verification is used to ensure benchmark-quality data.
In practice, the verifier provides a strong starting set from which annotators can efficiently curate high-quality examples for \bench.

\begin{figure*}[!ht]
    \centering
    \resizebox{0.8\textwidth}{!}{%
        \begin{minipage}{\textwidth}
            \centering
            \begin{subfigure}[b]{0.48\textwidth}
                \centering
                \includegraphics[width=\textwidth]{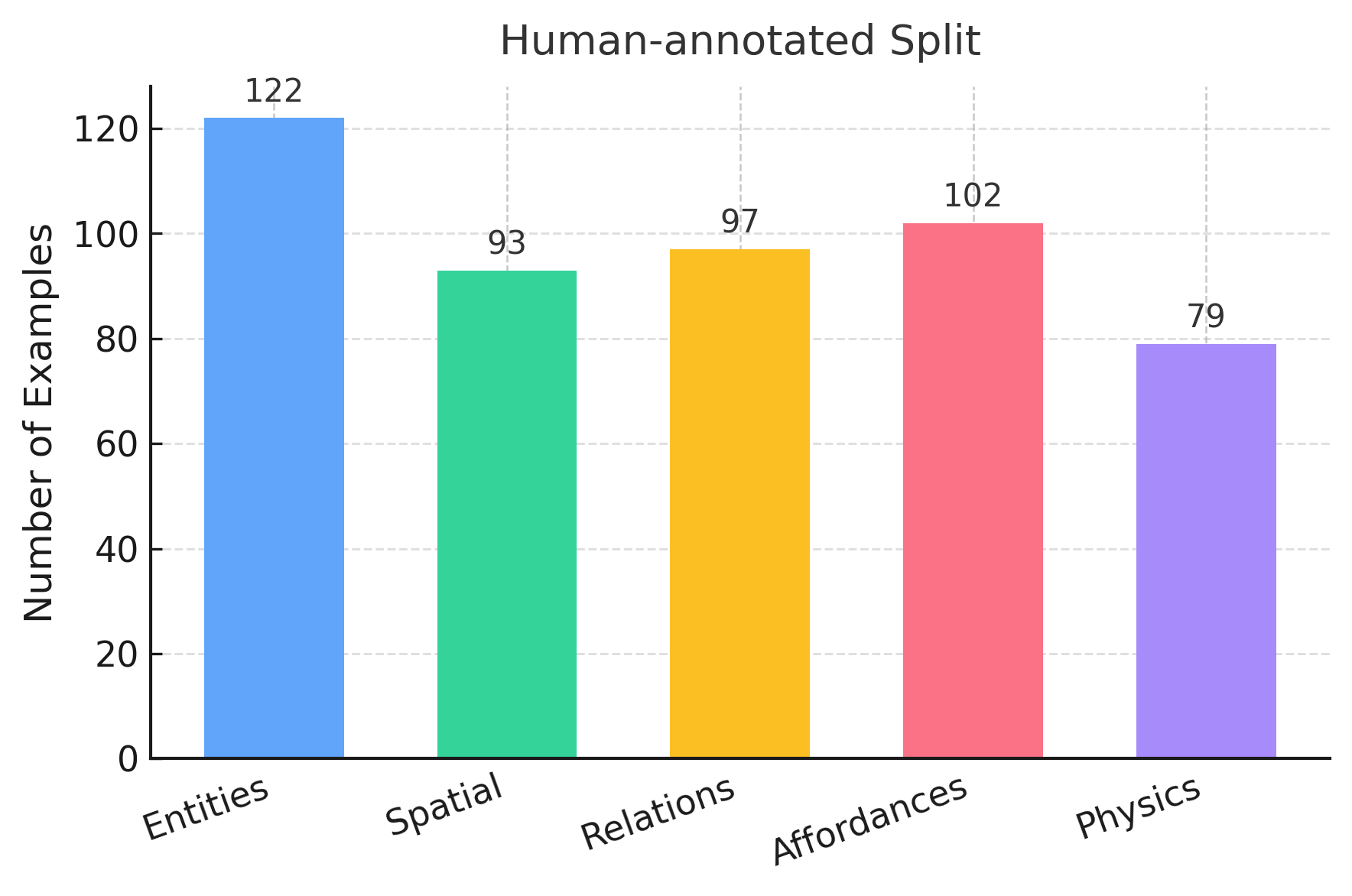}
                \caption{Human-annotated split}
                \label{fig:human-split}
            \end{subfigure}
            \hfill
            \begin{subfigure}[b]{0.48\textwidth}
                \centering
                \includegraphics[width=\textwidth]{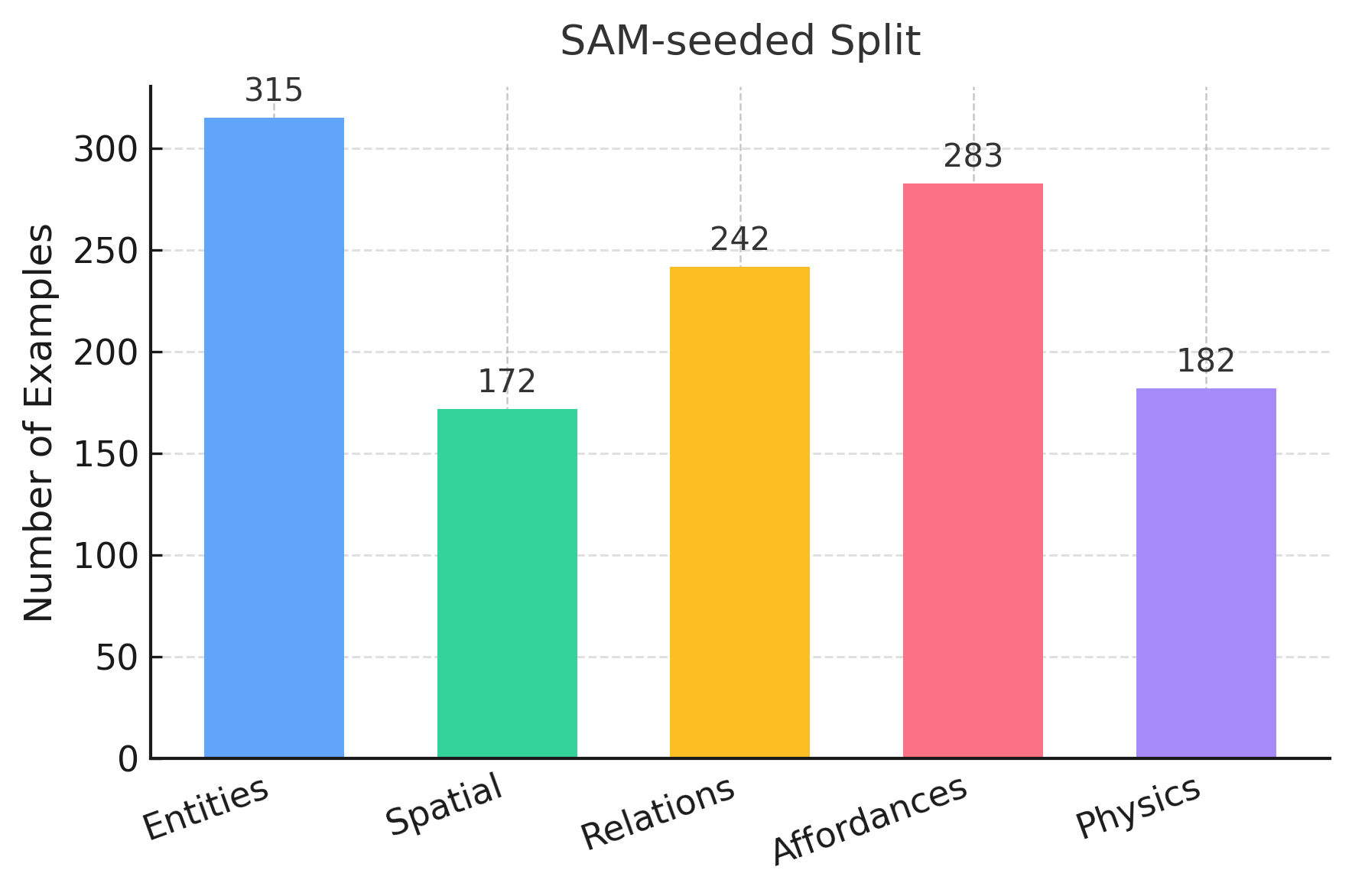}
                \caption{SAM-seeded split}
                \label{fig:sam-split}
            \end{subfigure}
        \end{minipage}
    }
    \caption{Distribution of examples per concept in the two splits of \bench.}
    \label{fig:benchmark-splits}
\end{figure*}
\begin{figure}[!htbp]
    \centering
    \includegraphics[width=\linewidth]{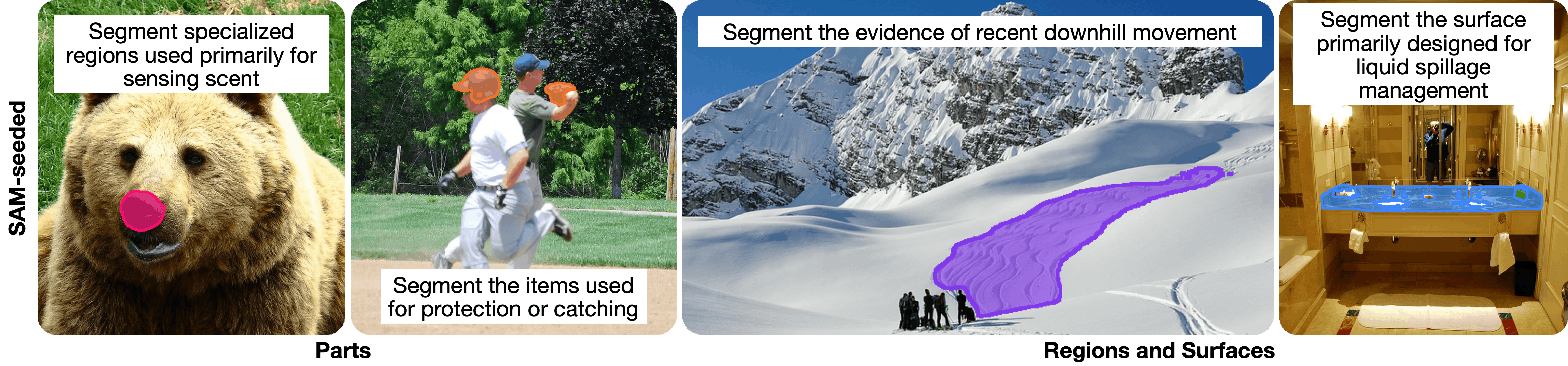}
    \caption{Examples of non-instance regions in \bench. The masks capture object parts, surfaces, and functional areas, demonstrating coverage of diverse region types beyond complete object instances.}
    \label{fig:supp-non-instance-examples}
\end{figure}
\subsection{Additional Statistics and Visualizations}
\label{sec:benchmark:stats}

In \cref{fig:benchmark-splits}, we show bar charts indicating the number of examples per concept family (entities, spatial \& layout, relations \& events, affordances \& functions, physics \& safety) for each split of \bench.
These statistics complement the distributional analysis in the main paper and confirm that all concept families are well represented.

\mypar{Region Type Diversity.}
Beyond whole-object instances, \bench includes diverse region types such as object parts, surfaces, and functional areas. The SAM-seeded split naturally incorporates these non-instance regions because SAM2 can generate masks for parts and surfaces in addition to complete objects. The human-annotated split further incorporates ``stuff'' regions from COCO-Panoptic, such as sky, grass, walls, and other amorphous areas. \cref{fig:supp-non-instance-examples} shows representative examples from the SAM-seeded split, including object parts, surfaces, and functional regions.

\subsection{Additional Qualitative Examples}
\label{sec:benchmark:additional-benchmark}

\begin{figure*}[!t]

\begin{center}

\includegraphics[width=0.9\linewidth]{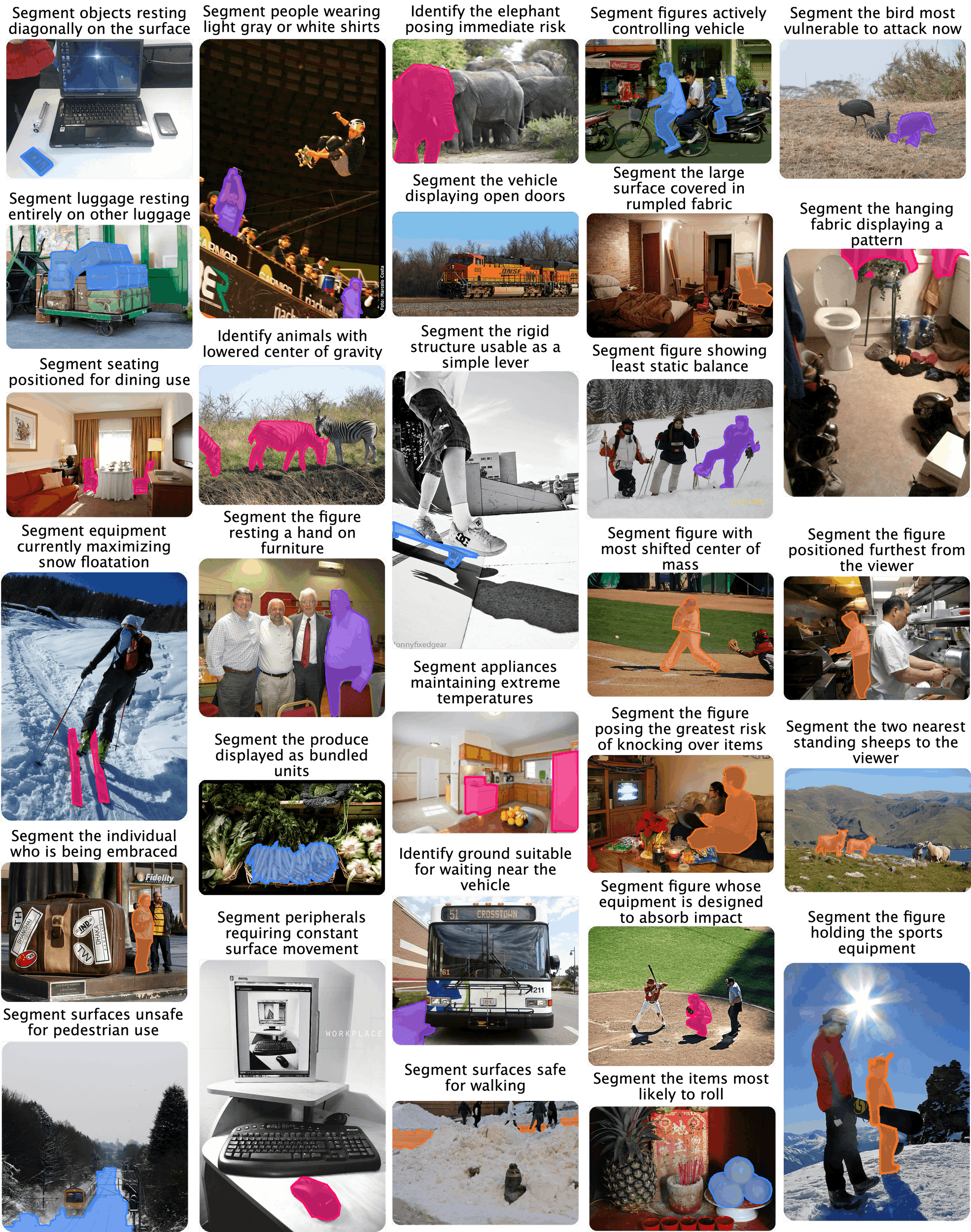} 

\end{center}

\vspace{-5mm}
\caption{\textbf{Additional qualitative examples from \bench.}
Each panel shows an input image, its conversational prompt, and the corresponding ground-truth mask overlaid.
Examples span all concept families (entities, spatial \& layout, relations \& events, affordances \& functions, physics \& safety) and belong to the human-annotated split.}

\label{fig:supp-benchmark-qualitative}
\vspace{-4mm}
\end{figure*}

We provide additional qualitative examples from \bench in \cref{fig:supp-benchmark-qualitative}, illustrating the diversity of conversational prompts and corresponding masks across concept families and splits.


\section{Implementation Details}
\label{sec:implementation}

\subsection{Architecture}
\label{sec:implementation:architecture}

\mypar{Prompt encoder.}
We use Qwen2.5-VL-3B~\cite{bai2025qwen2} as a frozen multimodal backbone: it takes both the RGB image and the conversational prompt as input. From the final hidden states we keep only positions corresponding to text tokens (padding, special tokens, and image placeholders are discarded). These token embeddings are layer-normalized and linearly projected to the SAM2 decoder width to form sparse language tokens. The EOS embedding is passed through a small MLP and broadcast as a $C \times H \times W$ dense bias map. Only the adapter projections and LoRA weights on top of Qwen are trained.

\mypar{Mask decoder.}
We use the SAM2~\cite{ravi2024sam} Hiera-L configuration (\texttt{sam2\_hiera\_l.yaml}). The image encoder is frozen. The mask decoder takes the sparse and dense language embeddings as prompt inputs. We supervise only the first output mask.

\subsection{Training Hyperparameters}
\label{sec:implementation:hyper}

We fine-tune the SAM2 mask decoder, SAM2 prompt encoder, and language adapter in two stages (pretraining and conversational post-training). Each stage is trained for $35\,000$ steps with AdamW and a cosine schedule with linear warmup. Images are resized so that the longer side is $1024$ pixels; masks are binarized and eroded with a $5 \times 5$ kernel (one iteration). We use a batch size of 6 with no gradient accumulation. The main optimization and model hyperparameters are summarized in~\cref{tab:train-hparams}.

\begin{table}[t]
\centering
\small
\setlength{\tabcolsep}{4pt}
\resizebox{\columnwidth}{!}{%
\begin{tabular}{@{}l|c@{}}
\toprule
Hyperparameter & Value \\
\midrule
Optimizer & AdamW \\
Learning rate ($\eta$) & $1\times10^{-4}$ \\
LR schedule & Warmup + cosine (min $10^{-6}$) \\
Weight decay & 0.05 \\
Batch size / grad.\ accum. & 6 / 1 \\
Steps (Stage 1, Stage 2) & $100\,000$, $90\,000$ \\
Image resolution & $1024 \times 1024$ (longer side) \\
LoRA rank and alpha ($r, \alpha$) & $16, 32$ \\
\bottomrule
\end{tabular}%
}
\caption{Training hyperparameters for \method.}
\label{tab:train-hparams}
\end{table}

\section{Additional Quantitative Results}
\label{sec:additional_results}

In this section, we present additional quantitative comparisons between \model and existing baselines.

\newcolumntype{A}{>{\columncolor{black!5}}c} 
\begin{table*}[t]
\centering
\footnotesize
\setlength{\tabcolsep}{4pt}
\renewcommand{\arraystretch}{1.15}
\resizebox{\linewidth}{!}{
\begin{tabular}{l l A *{2}{c} A *{2}{c} A *{2}{c} c A c c}
\toprule
\multirow{2}{*}{Model} & \multirow{2}{*}{Prompt Encoder} & \multicolumn{3}{c}{RefCOCO} & \multicolumn{3}{c}{RefCOCO+} & \multicolumn{3}{c}{RefCOCOg} & \multicolumn{4}{c}{ReasonSeg} \\
\cmidrule(lr){3-5}\cmidrule(lr){6-8}\cmidrule(lr){9-11}\cmidrule(lr){12-15}
& & val & testA & testB & val & testA & testB & val(U) & test(U) & val(G) & val & test & test(short) & test(long) \\
\midrule

LISA      & LLaVA 7B      & 74.9 & 79.1 & 72.3 & 65.1 & 70.8 & 58.1 & 67.9 & 70.6 & -- & 44.4 & 36.8  & 37.6  & 36.6 \\
LISA$^\star$      & LLaVA 7B        & -- & -- & -- & -- & -- & -- & -- & -- & -- & 52.9 & 47.3  & 40.6  & 49.4 \\
LISA      & LLaVA 13B       & -- & -- & -- & -- & -- & -- & -- & -- & -- & 48.9 & 44.8  & 39.9  & 46.4 \\
LISA$^\star$      & LLaVA 13B       & -- & -- & -- & -- & -- & -- & -- & -- & -- & 56.2 & 51.7  & 44.3  & 54.0 \\
LISA$^\star$      & Llama2 13B     & -- & -- & -- & -- & -- & -- & -- & -- & -- & 60.0 & 51.5  & 43.9  & 54.0 \\
LISA      & LLaVA1.5 7B     & -- & -- & -- & -- & -- & -- & -- & -- & -- & 53.6 & 48.8  & 48.3  & 49.2 \\
LISA$^\star$      & LLaVA1.5 7B     & -- & -- & -- & -- & -- & -- & -- & -- & -- & 61.3 & 55.6  & 48.3  & 57.9 \\
LISA      & LLaVA1.5 13B    & -- & -- & -- & -- & -- & -- & -- & -- & -- & 57.7 & 53.8  & 50.8  & 54.7 \\
LISA$^\star$      & LLaVA1.5 13B    & -- & -- & -- & -- & -- & -- & -- & -- & -- & 65.0 & 61.3  & 55.4  & 63.2 \\
SEEM                 & --              & -- & -- & -- & -- & -- & -- & -- & 75.1 & -- & 25.5 & 24.3  & 20.1  & 25.6 \\
Grounded SAM         & --              & -- & -- & -- & -- & -- & -- & -- & -- & -- & 26.0 & 21.3  & 17.8  & 22.4 \\
OVSeg                & --              & -- & -- & -- & -- & -- & -- & -- & -- & -- & 28.5 & 26.1  & 18.0  & 28.7 \\
Seg-Zero             & Qwen2.5-VL 3B   & -- & -- & -- & -- & -- & -- & -- & -- & -- & 58.2 & 56.1  & -- & -- \\
Seg-Zero             & Qwen2.5-VL 7B   & -- & -- & -- & -- & -- & -- & -- & -- & -- & 62.6 & 57.5  & -- & -- \\
GSVA$^\star$      & Vicuna 13B           & 79.2 & 81.7 & 77.1 & 70.3 & 73.8 & 63.6 & 75.7 & 77.0 & -- & -- & -- & -- & -- \\
GLaMM     & Vicuna 7B     & 79.5 & 83.2 & 76.9 & 72.6 & 78.7 & 64.6 & 74.2 & 74.9 & -- & 47.4 & --  & -- & -- \\
SAM4MLLM  & Qwen-VL 7B    & -- & -- & -- & -- & -- & -- & -- & -- & -- & 46.7 & -- & -- & -- \\
SAM4MLLM  & LLaVA1.6 7B   & 79.6 & 82.8 & 76.1 & 73.5 & 77.8 & 65.8 & 74.5 & 75.6 & -- & -- & -- & -- & -- \\
SAM4MLLM  & LLaVA1.6 8B   & 79.8 & 82.7 & 74.7 & 74.6 & 80.0 & 67.2 & 75.5 & 76.4 & -- &  58.4 & -- & -- & -- \\
GLEE      & CLIP          & 79.5 & -- & -- & 68.3 & -- & -- & 70.6 & -- & -- & -- & -- & -- & -- \\
GLEE      & CLIP           & 80.0 & -- & -- & 69.6 & -- & -- & 72.9 & -- & -- & -- & -- & -- & -- \\
DETRIS-L    & CLIP      & 81.0 & 81.9 & 79.0 & 75.2 & 78.6 & 70.2 & 74.6 & 75.3 & -- & -- & -- & -- & -- \\
UniLSeg-20   & CLIP ViT-B/16    & 80.5 & 81.8 & 78.4 & 72.7 & 77.0 & 67.0 & 78.4 & 79.5 & -- & -- & -- & -- & -- \\
UniLSeg-100   & CLIP ViT-B/16   & 81.7 & 83.2 & 79.9 & 73.2 & 78.3 & 68.2 & 79.3 & 80.5 & -- & -- & -- & -- & -- \\
PSALM     & Phi1.5 1.3B   & 83.6 & 84.7 & 81.6 & 72.9 & 75.5 & 70.1 & 73.8 & 74.4 & -- & -- & -- & -- & -- \\
EVF-SAM$^\dagger$   & BEIT-3-Large            & 82.1 & 83.7 & 80.0 & 75.2 & 78.3 & 70.1 & 76.8 & 77.4 & -- & -- & -- & -- & -- \\
EVF-SAM$^\ddagger$   & BEIT-3-Large    & 82.4 & 84.2 & 80.2 & 76.5 & 80.0 & 71.9 & 78.2 & 78.3 & -- & -- & -- & -- & -- \\
RICE      & Qwen2.5-7B    & 83.5 & 85.3 & 81.7 & 79.4 & 82.8 & 75.4 & 79.8 & 80.4 & -- & -- & -- & -- & -- \\
MLCD-seg  & Qwen2.5-7B    & 83.6 & 85.3 & 81.5 & 79.4 & 82.9 & 75.6 & 79.7 & 80.5 & -- & -- & -- & -- & -- \\
HyperSeg  & Phi2 2.7B     & 84.8 & 85.7 & 83.4 & 79.0 & 83.5 & 75.2 & 79.4 & 78.9 & -- & -- & -- & -- & -- \\
HyperSeg  & Phi2 3B         & -- & -- & -- & -- & -- & -- & -- & -- & -- & 59.2 & -- & -- & -- \\
Gemini Seg & Gemini2.5 Flash & -- & -- & -- & -- & -- & -- & -- & -- & -- & 28.3 & 30.6  & 16.5  & 35.0 \\
X-SAM     & Phi3 3.8B     & 85.1 & 87.1 & 83.4 & 78.0 & 81.0 & 74.4 & 83.8 & 83.9 & -- & 56.6 & 57.8  & 47.7  & 56.0 \\
RSVP       & LLaVA1.6 7B     & -- & -- & -- & -- & -- & -- & -- & -- & -- & 59.2 & 56.9  & 47.9  & 58.4 \\
RSVP       & Qwen2-VL 7B     & -- & -- & -- & -- & -- & -- & -- & -- & -- & 58.6 & 56.1  & 48.5  & 57.1 \\
RSVP       & Gemini1.5-Flash & -- & -- & -- & -- & -- & -- & -- & -- & -- & 56.9 & 57.1  & 47.3  & 60.2 \\
RSVP       & GPT-4o          & -- & -- & -- & -- & -- & -- & -- & -- & -- & 64.7 & 60.3  & 55.4  & 61.9 \\

\midrule
\model (Base) & Qwen2.5-VL 3B     &  78.4 & 80.8 & 75.8 & 72.5 & 77.7 & 66.4 & 75.1 & 74.7 & 74.7 & 51.1 & 48.3 & 47.2 & 48.6 \\
\model & Qwen2.5-VL 3B     &  78.2 & 80.3 & 74.7 & 72.0 & 77.5 & 66.3 & 74.1 & 73.9 & 73.7 & 56.4 & 52.2 & 53.8 & 51.7 \\
\model & Qwen2.5-VL 7B     &  79.4 & 81.6 & 76.4 & 74.3 & 79.1 & 69.2 & 74.9 & 75.5 & 75.0 & 61.9 & 57.0 & 54.2 & 57.9 \\

\bottomrule
\end{tabular}
}
\caption{\textbf{Referring expression segmentation (gIoU, \%).} \method is competitive on RefCOCO/+/g and achieves strong zero-shot performance on ReasonSeg, surpassing methods fine-tuned on ReasonSeg ($^\star$). $^\dagger$ trained on RefCOCO only; $^\ddagger$ on RefCOCO plus additional datasets (Objects365, PACO-LVIS, PASCAL-Part, etc).}
\label{tab:supp:referbench-results-more-baselines}
\end{table*}

\subsection{Cumulative IoU (cIoU)}
\label{sec:additional_results:ciou}

In the main paper, we reported gIoU performance of \model on the RefCOCO/+/g and ReasonSeg benchmarks.
In~\cref{tab:supp:referbench-results-ciou}, we report the corresponding cumulative IoU (cIoU) for \model on the same benchmarks, complementing the gIoU results.
In~\cref{tab:supp:ciou_convseg_results}, we additionally report cIoU performance of \model on \bench.

\newcolumntype{A}{>{\columncolor{black!5}}c} 

\begin{table*}[!ht]
\centering
\scriptsize
\setlength{\tabcolsep}{3pt}
\renewcommand{\arraystretch}{1.12}
\resizebox{\linewidth}{!}{
\begin{tabular}{l l A *{5}{c} A *{5}{c}}
\toprule
\multirow{2}{*}{Model} & \multirow{2}{*}{Prompt Encoder}
& \multicolumn{6}{c}{SAM-seeded (cIoU)} 
& \multicolumn{6}{c}{Human-annotated (cIoU)} \\
& & All & Ent. & Spat. & Rel. & Aff. & Phys. & All & Ent. & Spat. & Rel. & Aff. & Phys. \\
\midrule





\model     & Qwen2.5-VL 3B     &  70.8 & 76.7 & 73.6 & 75.5 & 65.7 & 57.7 & 67.8 & 73.4 & 65.5 & 67.3 & 68.0 & 59.2 \\
\model     & Qwen2.5-VL 7B     &  72.9 & 78.7 & 71.9 & 78.2 & 68.0 & 62.3 & 67.5 & 66.9 & 72.3 & 66.1 & 68.3 & 64.6 \\
\bottomrule
\end{tabular}
}
\vspace{-2mm}
\caption{\textbf{\bench benchmark results (cIoU, \%).}
Each subset reports performance across the five concept categories -- \uline{Ent}ities, \uline{Spat}ial, \uline{Rel}ations, \uline{Aff}ordances, and \uline{Phys}ics \& Safety -- and summarizes across all (\emph{All}).}
\vspace{-3mm}
\label{tab:supp:ciou_convseg_results}
\end{table*}

\newcolumntype{A}{>{\columncolor{black!5}}c} 
\begin{table*}[htbp]
\centering
\footnotesize
\setlength{\tabcolsep}{4pt}
\renewcommand{\arraystretch}{1.15}
\resizebox{\linewidth}{!}{
\begin{tabular}{l l A *{2}{c} A *{2}{c} A *{2}{c} c A c c}
\toprule
\multirow{2}{*}{Model} & \multirow{2}{*}{Prompt Encoder} & \multicolumn{3}{c}{RefCOCO} & \multicolumn{3}{c}{RefCOCO+} & \multicolumn{3}{c}{RefCOCOg} & \multicolumn{4}{c}{ReasonSeg} \\
\cmidrule(lr){3-5}\cmidrule(lr){6-8}\cmidrule(lr){9-11}\cmidrule(lr){12-15}
& & val & testA & testB & val & testA & testB & val(U) & test(U) & val(G) & val & test & test(short) & test(long) \\
\midrule

\model & Qwen2.5-VL 3B     & 77.6 & 79.4 & 73.9 & 71.7 & 76.1 & 65.6 & 74.1 & 73.7 & 75.5 & 64.0 & 56.6 & 51.6 & 58.0 \\
\model & Qwen2.5-VL 7B     & 79.0 & 81.0 & 75.9 & 74.4 & 78.1 & 68.6 & 75.3 & 75.7 & 76.8 & 64.3 & 60.4 & 53.5 & 62.2 \\

\bottomrule
\end{tabular}
}
\vspace{-2mm}
\caption{\textbf{Referring expression segmentation (cIoU, \%).} \method is competitive on RefCOCO/+/g and shows strong zero-shot performance on ReasonSeg.}

\vspace{-4mm}
\label{tab:supp:referbench-results-ciou}
\end{table*}

\newcolumntype{A}{>{\columncolor{black!5}}c} 

\begin{table*}[!ht]
\centering
\scriptsize
\setlength{\tabcolsep}{3pt}
\renewcommand{\arraystretch}{1.12}
\resizebox{\linewidth}{!}{
\begin{tabular}{l l A *{5}{c} A *{5}{c}}
\toprule
\multirow{2}{*}{Model} & \multirow{2}{*}{Prompt Encoder}
& \multicolumn{6}{c}{SAM-seeded (cIoU)} 
& \multicolumn{6}{c}{Human-annotated (cIoU)} \\
& & All & Ent. & Spat. & Rel. & Aff. & Phys. & All & Ent. & Spat. & Rel. & Aff. & Phys. \\
\midrule
SAM3  & Perception Encoder & 39.7 & 47.5 & 40.2 & 44.1 & 35.7 & 25.9 & 35.4 & 45.8 & 27.0 & 32.6 & 32.5 & 36.6 \\
\midrule
\model     & Qwen2.5-VL 3B     &   \textbf{70.8} & 74.0 & 70.9 & 74.1 & 68.7 & 64.2 & \textbf{67.4} & 71.6 & 68.7 & 67.0 & 64.4 & 63.8 \\
\model     & Qwen2.5-VL 7B     &   \textbf{72.4 }& 76.1 & 71.1 & 77.5 & 70.4 & 63.7 & \textbf{67.9} & 70.0 & 71.5 & 69.3 & 63.5 & 64.0 \\
\bottomrule
\end{tabular}
}
\vspace{-2mm}
\caption{\textbf{Comparison with SAM3 on \bench (gIoU \%).}
Each subset reports performance across the five concept categories -- \uline{Ent}ities, \uline{Spat}ial, \uline{Rel}ations, \uline{Aff}ordances, and \uline{Phys}ics \& Safety -- and summarizes across all (\emph{All}).}
\vspace{-3mm}
\label{tab:supp:convseg_results_sam3}
\end{table*}

\subsection{Additional Baselines}
\label{sec:additional_results:baselines}

We extend Table~2 of the main paper to include additional baselines and report gIoU performance in~\cref{tab:supp:referbench-results-more-baselines}.
This expanded comparison situates \model among a broader set of contemporary referring and reasoning segmentation approaches and provides a more complete view of the current landscape.

\mypar{Comparison with SAM3.} The recent work, SAM3~\cite{carion2025sam}, is a new variant of SAM that supports natural language promptable segmentation. We evaluate SAM3 on \bench and report results in~\cref{tab:supp:convseg_results_sam3}. SAM3 achieves 39.7\% gIoU on the SAM-seeded split and 35.4\% on the human-annotated split, substantially lower than \model (70.8\% and 67.4\% respectively with the 3B backbone). This demonstrates that our conversational training approach and concept-driven data engine provide significant gains for abstract reasoning in conversational image segmentation.






%


\end{document}